\documentclass{article}

\PassOptionsToPackage{numbers, compress}{natbib}

\usepackage[preprint]{neurips_2026}

\usepackage{microtype}
\usepackage{graphicx}
\usepackage{subcaption}
\usepackage{booktabs} %

\usepackage[utf8]{inputenc} %
\usepackage[T1]{fontenc}    %

\usepackage[colorlinks=true, linkcolor={sb_blue}, citecolor={sb_blue}, filecolor={sb_blue}, urlcolor={sb_blue}]{hyperref}
\usepackage{url}            %
\usepackage{booktabs}       %
\usepackage{amsfonts}       %
\usepackage{nicefrac}       %
\usepackage{microtype}      %
\usepackage[dvipsnames]{xcolor}         %
\definecolor{sb_blue}{rgb}{0.0, 0.5, 0.75}
\usepackage{wrapfig}
\usepackage{multirow}
\usepackage[singlelinecheck=false]{caption}
\usepackage{subcaption}

\usepackage{listings}
\usepackage{soul} 
\usepackage{algorithm}
\usepackage{algorithmic}
\usepackage{graphicx}
\usepackage{amsmath}
\usepackage{cleveref}

\usepackage{threeparttable}

\usepackage{tikz}
\usepackage{pgfplots}
\usepackage{comment}
\definecolor{tab9}{HTML}{9C755F}
\definecolor{spec1}{HTML}{9e0142}
\definecolor{spec2}{HTML}{d53e4f}
\definecolor{spec4}{HTML}{fdae61}
\definecolor{spec5}{HTML}{fee08b}
\definecolor{spec6}{HTML}{e6f598}
\definecolor{spec7}{HTML}{abdda4}
\definecolor{spec8}{HTML}{66c2a5}
\definecolor{spec9}{HTML}{3288bd}
\definecolor{spec10}{HTML}{5e4fa2}
\definecolor{colorenglish}{HTML}{FF7F0E}
\definecolor{colorfinnish}{HTML}{1F77B4}
\colorlet{LightBlue}{CornflowerBlue!40!}

\def\linethickness{1mm}

\def\roundness{8mm}

\usepackage{chemfig}

\def\colorcompone{spec1}
\def\colorcomptwo{spec8}
\def\colorenglish{colorenglish}
\def\colorfinnish{colorfinnish}
\def\latentcolor{spec4}
\def\trapezoidcolor{ProcessBlue}
\def\pixelspacecolor{spec2}
\def\latentspacecolor{OliveGreen}
\def\latentspacecolor{spec9}

\def\condcolor{spec4}
\def\classcolor{tab9}

\usepackage{mathtools}
\newcommand{\cblock}[1]{\textcolor{#1}{\rule{2.5mm}{2.5mm}}}

\usetikzlibrary{positioning,calc,patterns,angles,arrows, decorations.markings,shapes,arrows.meta,fit}
\lstdefinestyle{graybox}{
  backgroundcolor=\color{gray!20}, %
  frame=single, %
  basicstyle=\ttfamily, %
  breaklines=true, %
}

\usepackage{amsmath}
\usepackage{amssymb}
\usepackage{mathtools}
\usepackage{amsthm}

\theoremstyle{plain}

\theoremstyle{definition}

\theoremstyle{remark}

\title{EMoE: Training-Free Expert Disagreement for Uncertainty-Aware Text-to-Image Diffusion}

\author{Lucas Berry\textsuperscript{1} \quad
  Axel Brando\textsuperscript{2} \quad
  Wei-Di Chang\textsuperscript{1} \\
  \textbf{Juan Camilo Gamboa Higuera\textsuperscript{3}} \quad
  \textbf{David Meger\textsuperscript{1}} \\
  \vspace{0.5em} \\
  \textsuperscript{1}McGill University \\
  \textsuperscript{2}Barcelona Supercomputing Center (BSC) \\
  \textsuperscript{3}Ideogram AI
}

\begin{document}

\maketitle

\begin{abstract}
  Large text-to-image diffusion models rarely expose reliable signals of when a prompt is likely to produce a poorly aligned generation, especially when training data is undisclosed. We study whether expert disagreement inside pre-trained mixture-of-experts (MoE) diffusion models can serve as a reliable estimate for epistemic uncertainty. We introduce EMoE, a training-free method that separates expert-specific computation paths at an early MoE layer, uses the same initial noise across paths, and measures variance among their latent representations after the first denoising step. This provides an uncertainty-aware prompt signal before full image generation, without auxiliary networks or training diffusion ensembles. On COCO and CC3M, EMoE ranks prompts by text-image alignment quality metrics more consistently than diffusion-specific and router-based baselines. We further apply EMoE to multilingual prompts and find systematic language-dependent differences in disagreement and generation quality, including shared-vocabulary effects. These results position EMoE as a practical diagnostic tool for prompt risk, model coverage, and bias analysis in MoE text-to-image diffusion models.
\end{abstract}

\section{Introduction} \label{sec:intro}

Text-to-image diffusion models have advanced rapidly, enabling faster generation \citep{song2020denoising, liu2023instaflow, yin2024onestep}, higher visual fidelity \citep{dhariwal2021diffusion, nichol2022glide, rombach2022high}, and video synthesis \citep{ho2022video, khachatryan2023text2video, bar2024lumiere}. 
Despite this progress, these models remain opaque: they provide limited insight into their uncertainty, training-data coverage, or prompts likely to produce unreliable generations \citep{berryshedding, chan2024hyper}. This is especially problematic when training data is incompletely documented, making failure modes difficult to anticipate before costly image generation.

A principled way to probe these limitations is through epistemic uncertainty, which captures uncertainty arising from insufficient or unrepresentative training data and is, in principle, reducible with additional data \citep{hora1996aleatory,der2009aleatory,hullermeier2021aleatoric}. Unlike aleatoric uncertainty, which reflects irreducible noise, epistemic uncertainty provides a direct signal of where a model is extrapolating beyond its training distribution. In generative settings, however, estimating epistemic uncertainty remains challenging due to the high dimensionality of outputs and the computational cost of standard ensemble-based approaches.

We introduce Epistemic Mixture of Experts (EMoE), a zero-shot framework for estimating epistemic uncertainty in the latent space of text-conditioned diffusion models. EMoE leverages pre-trained mixture-of-experts (MoE) architectures by separating expert-specific computation paths at an early MoE layer and measuring their disagreement inside the denoiser. By fixing the initial noise across expert paths, EMoE isolates uncertainty due to expert disagreement rather than stochastic sampling variation. This yields an early prompt-level uncertainty estimate before full image generation, enabling applications such as abstention or filtering in large-scale generative systems \citep{henderson2020towards, song2024high}.

\begin{figure*}[t]
    \centering
    \resizebox{\textwidth}{!}{\input{./figures/paper/fig_1_version_3.tex}}
    \caption{EMoE estimates different uncertainty levels for the same prompt expressed in English and Finnish. The English prompt ``Two teddy bears are sitting together in the grass.'' yields lower uncertainty, while the Finnish translation ``Kaksi nallekarhua istuu yhdessä nurmikolla.'' yields higher uncertainty. This illustrates how latent-space expert disagreement can reveal language-dependent reliability gaps.}
    \label{fig:president_sauce}
    \vspace{-.5cm}
\end{figure*}

EMoE also provides a practical tool for analyzing biases arising from uneven data coverage. \autoref{fig:president_sauce} shows that an English prompt yields low epistemic uncertainty, whereas its Finnish translation produces substantially higher uncertainty. This suggests reduced model confidence for underrepresented language inputs and highlights how latent-space epistemic uncertainty can diagnose data imbalance.

We evaluate EMoE on COCO and CC3M \citep{lin2014microsoft, sharma2018conceptual}, showing that its uncertainty estimates align with downstream text-image alignment quality metrics. We further apply EMoE to multilingual prompts across 25 languages, where elevated uncertainty identifies underrepresented and out-of-distribution language inputs even without access to training data. Our contributions are:
\begin{itemize}
    \item We introduce EMoE, a zero-shot framework for estimating epistemic uncertainty in text-to-image diffusion models by disentangling MoE representations in latent space.
    \item We demonstrate that EMoE uncertainty aligns with downstream text-image alignment and image quality metrics on COCO and CC3M.
    \item We apply EMoE to multilingual prompts across 25 languages and show that it reveals language-dependent reliability gaps and systematic biases.
    \item We provide ablations on ensemble size, denoising step, latent-space choice, and model architecture, confirming the robustness of EMoE.
\end{itemize}

\section{Background} \label{sec:background}

Diffusion models generate samples by iteratively removing Gaussian noise, forming a Markov chain. This probabilistic structure naturally supports uncertainty estimation \citep{hullermeier2021aleatoric}. In parallel, ensemble methods are widely used to estimate epistemic uncertainty through disagreement \citep{hoffmann2021deep}. Since MoE architectures contain multiple specialized components within a single model, they provide a natural foundation for estimating epistemic uncertainty from expert disagreement.

\subsection{Diffusion Models}

Let $x\in\mathbb{R}^{512\times512\times3}$ denote an image and $y$ its associated prompt. The objective is to estimate the conditional distribution $p(x|y)$. Due to the high-dimensional and multi-modal nature of image generation, many diffusion models operate in a lower-dimensional latent space learned by an autoencoder \citep{rombach2022high}. The autoencoder consists of an encoder $\mathcal{E}$, which maps images to latent representations, and a decoder $\mathcal{D}$, which maps latent representations back to image space. Thus, after encoding $x$ into $z_0=\mathcal{E}(x)$, the diffusion model estimates a latent conditional generation process, while $\mathcal{D}$ reconstructs the final image from the generated latent representation.

Diffusion models use a two-phase approach, consisting of a forward and a reverse process, to generate realistic images. In the forward phase, an initial image $x$ is encoded to $z_0$ and then gradually corrupted by adding Gaussian noise over $T$ steps, resulting in a sequence of noisy latent states $z_1, z_2, \ldots, z_T$. This process can be expressed as:
\begin{align*}
    &q(z_t|z_{t-1}) = \mathcal{N}(z_t;\sqrt{1-\beta_t}z_{t-1}, \beta_t{\bf I}) \\
    &q(z_{1:T}|z_0) = \prod_{t=1}^Tq(z_t|z_{t-1}),
\end{align*}
where $\beta_t\in(0,1)$, with $\beta_1 < \beta_2 < \dots < \beta_T$. This forward process draws inspiration from non-equilibrium statistical physics \citep{sohl2015deep}.

The reverse process aims to remove the noise and recover the original image, conditioned on text. This is achieved by estimating the conditional distribution $q(z_{t-1}|z_t,y)$ through a model $p_{\theta}$. The reverse process is defined as:
\begin{align*}
    &p_{\theta}(z_{0:T}|y)=p(z_T)\prod_{t=1}^Tp_{\theta}(z_{t-1}|z_t,y) \\
    &p_{\theta}(z_{t-1}|z_t,y)=\mathcal{N}(z_{t-1};\mu_{\theta}(z_t,t,y),\Sigma_t).
\end{align*}
where $p_{\theta}(z_{t-1}|z_t,y)$ is the denoising distribution, modeled as a Gaussian with learned mean $\mu_{\theta}(z_t,t,y)$ and predefined covariance $\Sigma_t$. Since the exact log-likelihood $\log(p_{\theta}(z_0|y))$ is intractable, diffusion models optimize an ELBO-based noise-prediction objective \citep{kingma2013auto, ho2020denoising}:
\begin{align*}
    L_{LDM} = \mathbb{E}_{z,\epsilon \sim \mathcal{N}(0,1),t,y}\left[||\epsilon-\epsilon_{\theta}(z_t,t,y)||^2_2\right],
\end{align*}
where $t \sim \mathrm{Unif}\{1,\ldots,T\}$, $\epsilon \sim \mathcal{N}(0,1)$, and $\epsilon_{\theta}$ predicts the noise for computing $\mu_{\theta}$.

\subsection{U-Nets}

U-Nets are a standard backbone for diffusion models because skip connections preserve both local and global information \citep{ronneberger2015u}. A U-Net consists of $\mathrm{down}$-blocks, a $\mathrm{mid}$-block, and $\mathrm{up}$-blocks. For the architectures used in our experiments, the downsampling path maps $z_t$ to $m^{\mathrm{pre}}_t\in\mathbb{R}^{1280\times8\times8}$, the mid-block refines it to $m^{\mathrm{post}}_t\in\mathbb{R}^{1280\times8\times8}$, and the upsampling path maps $m^{\mathrm{post}}_t$ to the next latent $z_{t-1}$. Thus, the U-Net models $\epsilon_{\theta}(z_t,t,y)$ by removing noise while preserving structure.

To condition on a prompt $y$, we encode it as $\tau_\theta(y)$ and inject it into the U-Net through cross-attention layers in the $\mathrm{down}$-, $\mathrm{mid}$-, and $\mathrm{up}$-blocks. The cross-attention operation is $\mathrm{Attention}(Q,K,V)=\mathrm{softmax}\!\left(QK^T/\sqrt{d}\right)V$, with
\begin{align*}
    Q = W_Q\phi_{\theta}(z_t), \quad K = W_K \tau_{\theta}(y), \quad V = W_V \tau_{\theta}(y).
\end{align*}
Here, $\phi_{\theta}(z_t)$ is the encoded noisy latent, $\tau_{\theta}(y)$ is the text representation, and $W_Q,W_K,W_V$ are learned projections \citep{vaswani2017attention}.

\subsection{Sparse Mixture of Experts}
\begin{figure}[t!]
    \vspace{0.2in}
    \centering
    \resizebox{\columnwidth}{!}{\input{./figures/paper/flow_chart.tex}}
    \caption{EMoE separates expert components in the first cross-attention layer in the first $\mathrm{down}$-block and processes each component separately as an independent computation path in the MoE pipeline. This results in $M$ distinct latent representations after the first denoising step. The figure illustrates an ensemble with two expert components, $\left(\cblock{spec8}\,\text{and}\,\cblock{spec1}\right)$.}
    \label{fig:flowchart}
    \vspace{-0.2in}
\end{figure}

MoE combines specialized models, or ``experts'', whose contributions are dynamically selected or weighted for each input \citep{jacobs1991adaptive}. Our pre-trained models contain sparse MoE layers in the U-Net's cross-attention and feed-forward blocks \citep{shazeer2017outrageously, fedus2022switch}. For $M$ experts, the $i$-th cross-attention expert has expert-specific projections
\begin{align}
    Q^i = W^i_Q \phi_{\theta}(z_t), \quad K^i = W^i_K \tau_{\theta}(y), \quad V^i = W^i_V \tau_{\theta}(y), \label{eq:expert_QKV}
\end{align}
where $W^i_Q$, $W^i_K$, and $W^i_V$ are learned projection matrices. A sparse MoE layer selects a subset of experts $\mathcal{S}\subseteq\{1,\ldots,M\}$ and combines them with routing weights $w^i$. For cross-attention, this gives
\begin{align}
Q = \sum_{i \in \mathcal{S}} w^i Q^i, \quad K = \sum_{i \in \mathcal{S}} w^i K^i, \quad V = \sum_{i \in \mathcal{S}} w^i V^i. \label{eq:expert_weighted_sum}
\end{align}
Feed-forward MoE layers aggregate expert outputs similarly \citep{lepikhin2020gshard}, enabling efficient scaling by activating only a subset of experts per input.

\section{Epistemic Mixture of Experts} \label{sec:EMoE}

Epistemic uncertainty arises when inputs are insufficiently represented in training data \citep{gruber2023sources, wang2024epistemic}. EMoE estimates this uncertainty in the denoiser's latent space by separating expert-specific computation paths in an MoE diffusion model. Unlike standard MoE architectures, which aggregate expert outputs, EMoE preserves expert-conditioned latent trajectories and measures their disagreement, analogous to ensemble-based uncertainty estimation \citep{lakshminarayanan2017simple}.

Here, disentanglement means separating latent trajectories induced by the same prompt and noise, so variation across paths reflects learned expert differences rather than diffusion sampling noise.

\subsection{Separation of Experts}

In sparse MoE diffusion models, expert contributions are normally combined at each MoE layer. EMoE instead separates experts at the first sparse MoE layer, located in the initial cross-attention layer of the first downsampling block. For expert $i$, we compute
\begin{align*}
CA^i = \mathrm{Attention}(Q^i, K^i, V^i),
\end{align*}
where $Q^i$, $K^i$, and $V^i$ are defined in \autoref{eq:expert_QKV}.

Each resulting branch then propagates independently through the rest of the U-Net. Later sparse MoE layers use the original routing and aggregation mechanism within each branch, so branches remain distinct because they start from different expert-conditioned activations. This yields $M$ latent representations during denoising rather than one aggregated representation, as shown in \autoref{fig:flowchart}.

Separating experts this early allows epistemic uncertainty to be estimated after one denoising step instead of all $T$ reverse steps, enabling abstention or early filtering before full image generation.

\subsection{Epistemic Uncertainty Estimation}

We connect EMoE to a standard decomposition of total uncertainty into aleatoric and epistemic components. Following variance-based decompositions used in ensemble and diffusion uncertainty estimation \citep{lakshminarayanan2017simple, chan2024hyper, depeweg2018decomposition}, total uncertainty can be written as
\begin{align}
\mathrm{TU}(y) = \mathrm{AU}(y) + \mathrm{EU}(y).
\end{align}
For a predictive model with parameters $\theta$, this follows from the law of total variance:
\begin{align}
\mathrm{Var}(x\mid y, \mathcal{D})
=
\underbrace{
\mathbb{E}_{\theta \sim p(\theta\mid\mathcal{D})}
\left[
\mathrm{Var}_{x\sim p(x\mid y,\theta)}
(x)
\right]
}_{\mathrm{AU}(y)}
+
\underbrace{
\mathrm{Var}_{\theta \sim p(\theta\mid\mathcal{D})}
\left[
\mathbb{E}_{x\sim p(x\mid y,\theta)}
(x)
\right]
}_{\mathrm{EU}(y)}.
\label{eq:tu_decomposition}
\end{align}
The aleatoric term captures output variability for a fixed model, while the epistemic term captures variability across model components. EMoE targets the epistemic component by measuring expert-conditioned latent disagreement while controlling the stochastic sampling source.

Rather than estimating uncertainty in pixel space after full generation, EMoE estimates epistemic uncertainty inside the denoiser. Specifically, we compute uncertainty after the U-Net mid-block using the expert-specific latent representations $m^{\mathrm{post}}_{T,i}$ obtained after the first denoising step.

Let $D_{\mathrm{mid}} = 1280 \times 8 \times 8$ denote the number of mid-block dimensions for the architectures used in our experiments, and let $m^{\mathrm{post}}_{T,i,d}$ denote dimension $d$ of the representation from expert path $i$. We define
\begin{align}
\mathrm{EU}(y)
=
\frac{1}{D_{\mathrm{mid}}}
\sum_{d=1}^{D_{\mathrm{mid}}}
\mathrm{Var}_{i=1,\ldots,M}
\left(
m^{\mathrm{post}}_{T,i,d}
\right).
\label{eq:emoe_estimator}
\end{align}
This estimator is the normalized trace covariance of the expert-conditioned mid-block representations. Equivalently, it is the average pairwise squared disagreement between expert paths, as shown in \autoref{sec:gp_analogy}. Thus, EMoE measures how separated the expert-conditioned representations are in the denoiser's latent space.

Since $m^{\mathrm{post}}_{T,i}$ is conditioned on prompt $y$, $\mathrm{EU}(y)$ gives a prompt-level estimate of epistemic uncertainty. Using the same initial noise $z_T$ across expert paths controls for sampling noise, so the variance in \autoref{eq:emoe_estimator} is induced by expert-specific latent representations rather than different stochastic generations. In \autoref{sec:gp_analogy}, we formalize this using a variance decomposition that separates expert disagreement at fixed noise from variation caused by changing the initial diffusion noise.

Higher values indicate stronger expert disagreement for the same prompt and initial noise. We interpret this controlled disagreement as epistemic uncertainty in the denoiser's latent space and empirically validate it through downstream alignment and distribution-shift experiments. Additional intuition from Gaussian Processes and ensemble-based uncertainty estimation is provided in \autoref{sec:gp_analogy}.

\subsection{Building MoE}
To capture epistemic uncertainty, expert components must be diverse enough to produce informative disagreement. Deep ensembles typically obtain diversity through bootstrapping or random initialization \citep{breiman2001random,lakshminarayanan2017simple}. EMoE instead uses pre-trained MoE checkpoints assembled from publicly available models on \href{https://huggingface.co/}{Hugging Face} and \href{https://civitai.com/}{Civit AI}, avoiding additional training. Since this diversity is inherited rather than optimized for uncertainty estimation, we test its sufficiency through ablations on ensemble size and closely related checkpoints. As shown in \autoref{sec:ablations}, even four checkpoints from the same model family provide useful disagreement. Training an equivalent diffusion ensemble from scratch would require roughly 150,000 GPU-hours per expert, costing approximately \$600,000 USD.

A gating module routes inputs to selected experts and assigns weights. Rather than training this module, we infer weights by comparing the prompt representation $\tau_{\theta}(y)$ with $\tau_{\theta}(\psi^i)$, where $\psi^i$ describes expert $i$'s strengths and weaknesses:
\begin{align*}
    \alpha^i = \tau_{\theta}(\psi^i)\cdot \tau_{\theta}(y), \qquad w^i = \mathrm{softmax}(\alpha^i).
\end{align*}
This enables dynamic expert selection without additional training. Further details are provided in \autoref{sec:gates_train} and \citet{goddard2024arcee}.

\section{Results} \label{sec:results}

We evaluate EMoE on COCO and CC3M \citep{lin2014microsoft, sharma2018conceptual} using modified \href{https://github.com/huggingface/diffusers}{diffusers} and \href{https://github.com/segmind/segmoe}{segmoe} libraries \citep{von-platen-etal-2022-diffusers,segmoe}. Unless otherwise stated, experiments use a single pre-trained four-expert SegMoE model (\autoref{sec:model_cards}). Because this model was not designed or trained for uncertainty estimation, it provides a neutral testbed and avoids selection bias from manually choosing experts.

For multilingual evaluation, we translate English COCO prompts using the Google translation API. This follows established multilingual vision-language practice: COCO-35L translates COCO captions into 35 languages using Google's translation API \citep{thapliyal2022crossmodal}. Translated COCO prompts provide a controlled way to approximately preserve semantic content while varying language, especially because COCO prompts are short, concrete, and caption-like.

To test whether EMoE's latent-space epistemic uncertainty predicts downstream reliability, we evaluate it in three settings: in-distribution English prompts, Finnish prompts as an out-of-distribution language setting, and multilingual prompts across 25 languages. We report CLIP score \citep{hessel2021clipscore} for text-image alignment, using the original English COCO prompt to score non-English generations against the same semantic content. Since CLIP-based evaluation can introduce biases, we discuss limitations in \autoref{sec:bias_clip} and corroborate trends with Image Reward \citep{schuhmann2022laion} and Aesthetic Score \citep{xu2024imagereward}.

\subsection{English Prompts} 
\label{sec:english_prompts}
\begin{wrapfigure}{r}{0.50\textwidth}
\vspace{-0.15in}
\centering

\includegraphics[width=\linewidth]{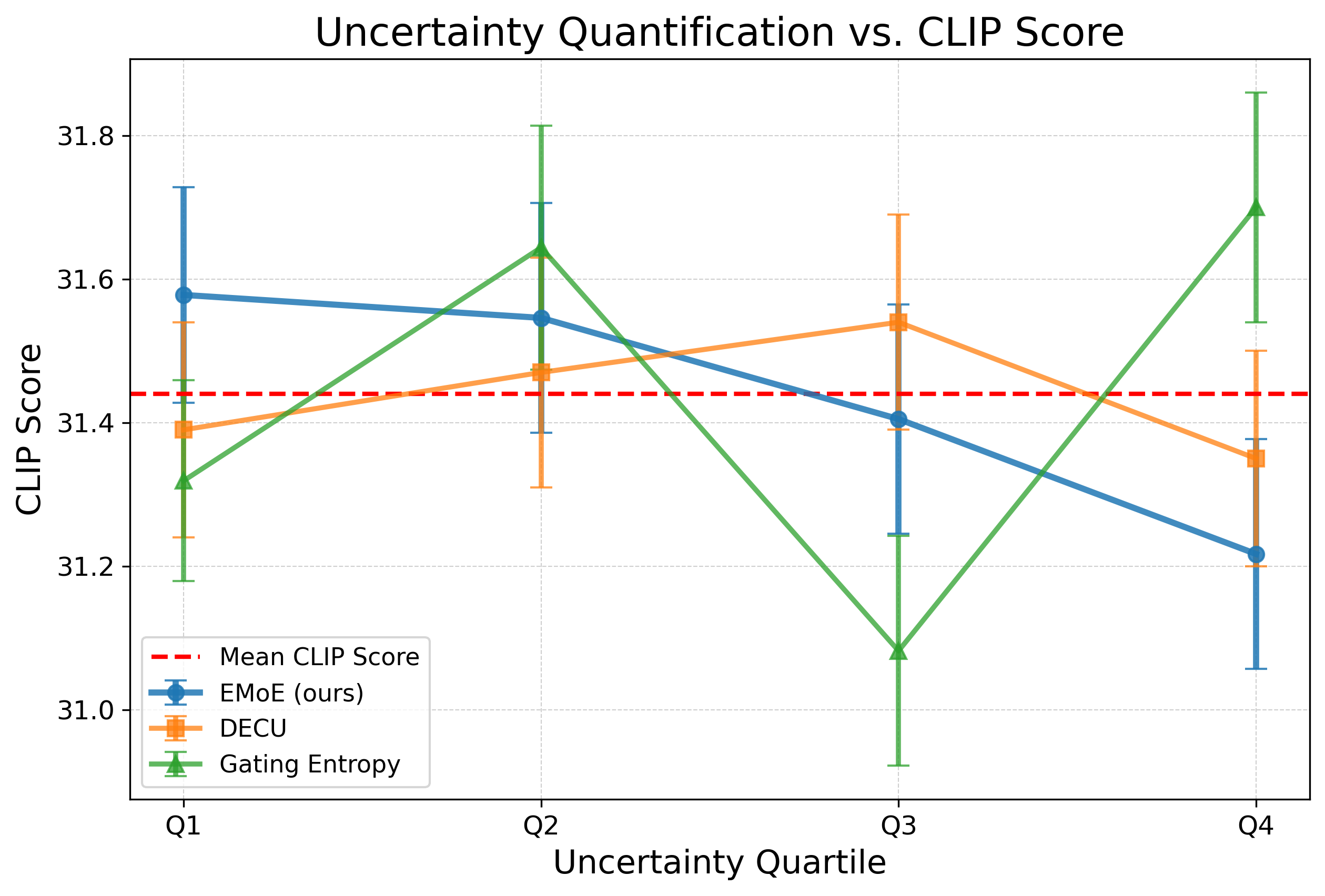}
\caption{CLIP score across uncertainty quartiles. EMoE assigns lower uncertainty to prompts with higher CLIP scores, unlike DECU and Gating Entropy. The red line indicates the mean CLIP score.}
\label{fig:english_clip_score}

\vspace{0.08in}

\captionof{table}{CLIP score, Aesthetic score, and Image Reward across EMoE uncertainty quartiles on 40k English prompts.}
\label{tab:other_scores}
\resizebox{\linewidth}{!}{%
\begin{tabular}{cccc}
\hline
\textbf{Quartile} & \textbf{CLIP $\uparrow$} & \textbf{Aesthetic $\uparrow$} & \textbf{Reward $\uparrow$} \\ \hline
Q1 & 31.578$\pm$0.15 & 5.763 & 0.292 \\ \hline
Q2 & 31.546$\pm$0.16 & 5.744 & 0.290 \\ \hline
Q3 & 31.405$\pm$0.16 & 5.733 & 0.273 \\ \hline
Q4 & 31.217$\pm$0.16 & 5.682 & 0.266 \\ \hline
\end{tabular}%
}

\vspace{-0.15in}
\end{wrapfigure}

We first test whether EMoE ranks in-distribution prompts by generation reliability. We sample 40,000 English COCO prompts, compute $\mathrm{EU}(y)$ using the same initial noise $z_T$ across expert paths, generate images, and divide the prompt-image pairs into uncertainty quartiles from Q1 (lowest) to Q4 (highest). For each quartile, we evaluate text-image alignment with CLIP score. As shown in \autoref{fig:english_clip_score}, EMoE produces a monotonic ordering: lower-uncertainty prompts consistently achieve higher CLIP scores, while higher-uncertainty prompts score lower. Although the absolute differences are modest, as expected for in-distribution COCO prompts, statistical tests and effect sizes confirm that the trend is meaningful (\autoref{sec:stat_analysis}). This pattern is not observed with DECU \citep{berryshedding}. As a lightweight baseline, we also report Gating Entropy, defined as the entropy of the MoE routing distribution over experts. Its non-monotonic trend suggests that router uncertainty alone is insufficient to capture prompt-level reliability.

The trend also holds for other metrics. \hyperref[tab:other_scores]{\Cref{tab:other_scores}} shows that Aesthetic Score and Image Reward \citep{schuhmann2022laion, xu2024imagereward} decrease as EMoE uncertainty increases, indicating that latent epistemic uncertainty aligns with both semantic alignment and reward-based quality. Prompt-length analyses are provided in \autoref{sec:additional}; the effect is statistically significant but small. Finally, the same trend appears on CC3M \citep{sharma2018conceptual} (\autoref{sec:additional_dataset}), supporting robustness beyond COCO.

\subsection{Finnish Prompts} \label{sec:finnish_prompts}
\begin{wrapfigure}{r}{0.5\textwidth}
\vspace{-0.55in}
\centering

\includegraphics[width=\linewidth]{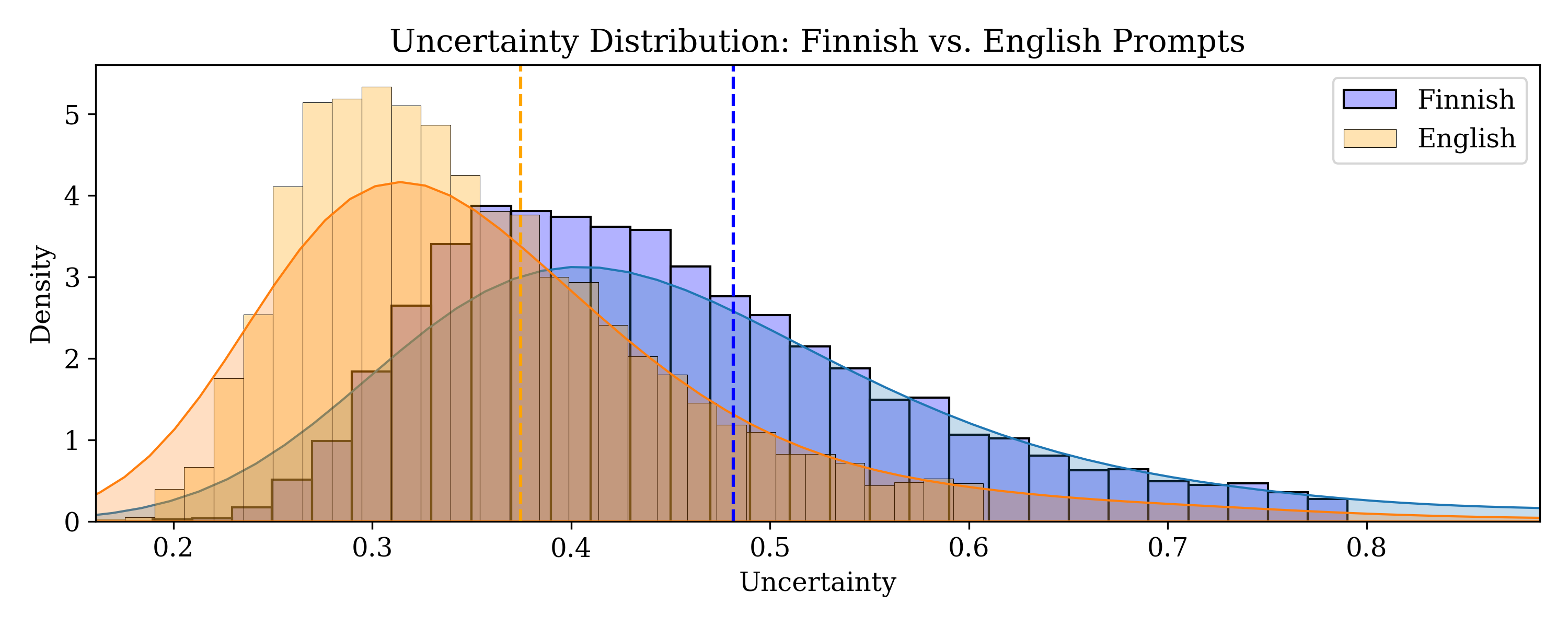}
\vspace{-0.1in}
\caption{Uncertainty distribution for Finnish and English prompts, showing higher uncertainty for Finnish prompts compared to English using EMoE.}
\label{fig:uncertainty_finnish_vs_english}

\vspace{0.05in}

\captionof{table}{CLIP scores and uncertainty statistics for Finnish and English prompts. Uncertainty scores are comparable across languages within each method, but not across methods.}
\label{tab:english_vs_finnish}
\resizebox{\linewidth}{!}{%
\begin{tabular}{lcccc}
\hline
\textbf{Language} & \textbf{CLIP Score $\uparrow$} & \textbf{EMoE} & \textbf{DECU} & \textbf{Gating Entropy} \\ \hline
Finnish & 16.41 & 0.48 $\pm$ 0.19 & 4.44 $\pm$ 5.22 & 0.05 $\pm$ 0.05 \\ \hline
English & 31.39 & 0.37 $\pm$ 0.14 & 4.04 $\pm$ 5.85 & 0.02 $\pm$ 0.05 \\ \hline
\end{tabular}%
}

\vspace{-0.15in}
\end{wrapfigure}

Next, we test whether EMoE detects language-level distribution shift. We translate 10,000 English COCO prompts into Finnish and compute uncertainty for both the original prompts and their translations. Since Finnish is less represented than English in many web-scale multimodal datasets, we expect Finnish prompts to induce higher epistemic uncertainty.

\autoref{fig:uncertainty_finnish_vs_english} compares Finnish and English uncertainty distributions for EMoE. Finnish prompts exhibit a clear shift toward higher uncertainty, indicating greater latent-space expert disagreement under language shift. This is also reflected in \hyperref[tab:english_vs_finnish]{\Cref{tab:english_vs_finnish}}, which reports CLIP scores alongside EMoE, DECU, and Gating Entropy uncertainty statistics. Finnish prompts have substantially lower CLIP scores and higher EMoE uncertainty than English prompts, while the other uncertainty estimates show weaker separation. Additional distribution plots for DECU and Gating Entropy are provided in \autoref{sec:additional}.

\begin{figure}[t!]
    \vskip 0.2in
    \centering
    \resizebox{\columnwidth}{!}{\input{./figures/paper/pizza.tex}}
    \caption{Qualitative comparison of image generation for a Finnish prompt with the word ``pizza'' and a random Finnish prompt. The English translation was not provided to the model.}
    \label{fig:pizza_bias}
    \vskip -0.2in
\end{figure}
\begin{wrapfigure}{r}{0.5\textwidth}
\vspace{-0.1in}
\centering

\captionof{table}{Finnish out-of-distribution detection using uncertainty scores. Finnish is treated as OOD and English as in-distribution.}
\label{tab:finnish_ood_auroc}
\resizebox{\linewidth}{!}{%
\begin{tabular}{lc}
\hline
\textbf{Uncertainty Signal} & \textbf{AUROC $\uparrow$} \\ \hline
EMoE & 0.739 \\ \hline
DECU & 0.571 \\ \hline
Gating Entropy & 0.543 \\ \hline
\end{tabular}%
}

\vspace{0.08in}

\captionof{table}{Proportion of prompts containing ``pizza'' in the lowest EMoE uncertainty quartile for Finnish and English prompts.}
\label{tab:pizza_bias}
\resizebox{0.78\linewidth}{!}{%
\begin{tabular}{cc}
\hline
\multirow{2}{*}{\textbf{Language}} 
& \textbf{Proportion of Prompts} \\
& \textbf{with ``pizza'' in Q1} \\ \hline
Finnish & 46.67\% \\ \hline
English & 21.54\% \\ \hline
\end{tabular}%
}

\vspace{0.08in}
    \centering
    \includegraphics[width=\linewidth]{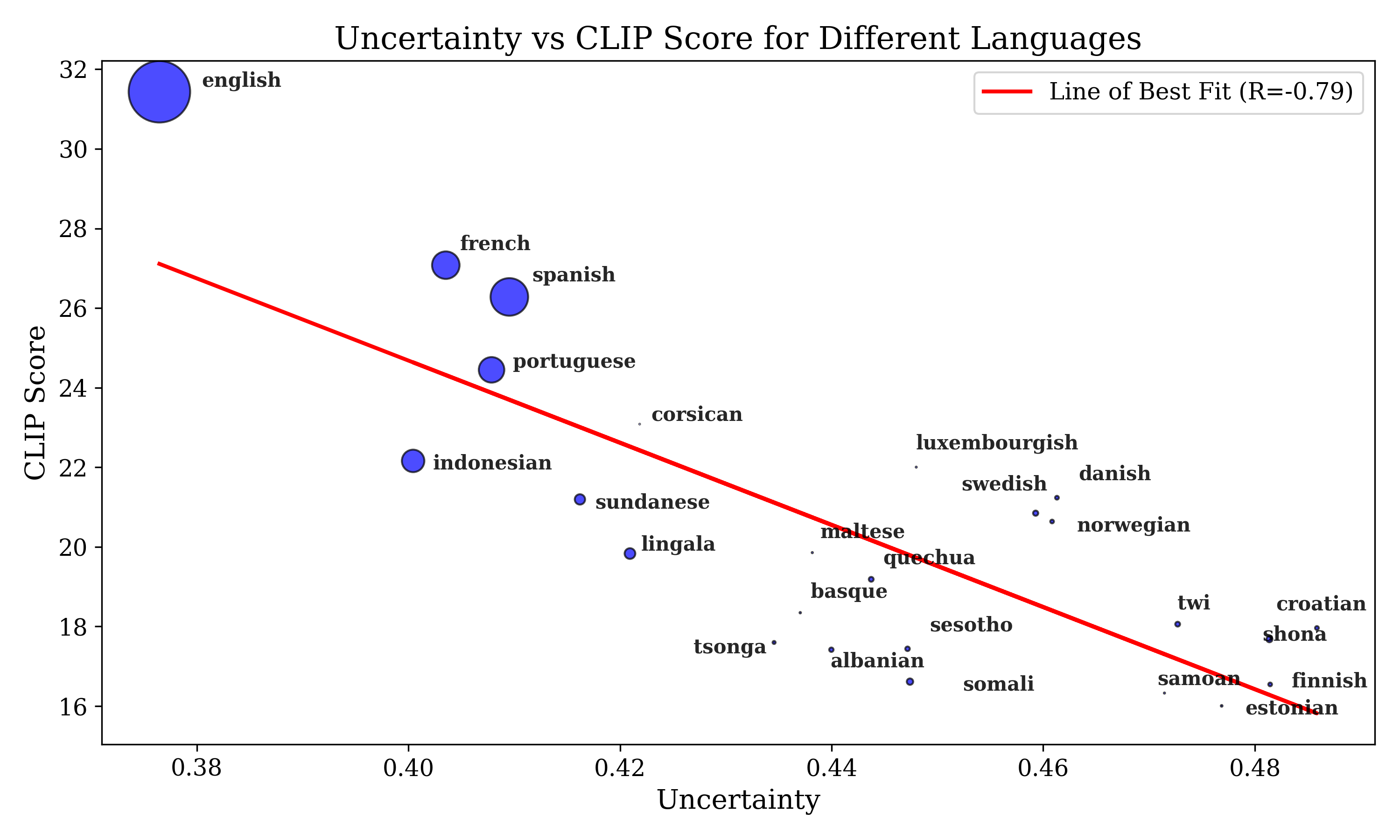}
    \captionof{figure}{Negative correlation between uncertainty and image quality across prompts translated into 25 different languages. EMoE demonstrates a strong negative correlation ($r=-0.79$) between uncertainty and CLIP score, with languages having more native speakers generally producing lower uncertainty and higher-quality images.}
    \label{fig:uncertainty_all_languages}
    \vskip -0.6in
\end{wrapfigure}
We quantify this separation in \hyperref[tab:finnish_ood_auroc]{\Cref{tab:finnish_ood_auroc}}, treating Finnish as out-of-distribution and English as in-distribution. EMoE achieves the strongest AUROC of $0.739$, outperforming DECU ($0.571$) and Gating Entropy ($0.543$). This provides a proof of concept that EMoE could support abstention or early filtering before full image generation. In contrast, the weaker DECU and Gating Entropy results suggest that direct diffusion latent variation and router uncertainty are less informative for this language-shift setting. This suggests that routing uncertainty and expert disagreement capture different phenomena: the router may be confident about which experts to activate, even when the separated expert-conditioned representations disagree in latent space.

We also use EMoE to probe lexical effects. Finnish prompts containing the shared-vocabulary word ``pizza'' often produce more text-aligned images than random Finnish prompts (\autoref{fig:pizza_bias}). EMoE captures this quantitatively: 46.67\% of Finnish ``pizza'' prompts fall into the lowest-uncertainty quartile, compared to 21.54\% for English prompts (\hyperref[tab:pizza_bias]{\Cref{tab:pizza_bias}}). This suggests that EMoE captures not only broad language shift, but also finer lexical effects that influence prompt grounding. A similar prompt-length trend for Finnish prompts is reported in \autoref{sec:additional} and supported by the ordered-trend analysis in \autoref{sec:stat_analysis}, but we treat prompt length as a contributing factor rather than a complete explanation for the uncertainty signal.

\subsection{Multilingual Prompts}  
\label{sec:all_languages_prompts}

To further test whether EMoE captures broader language-dependent reliability differences, we translate 1,000 COCO prompts into 23 additional languages and compute both EMoE uncertainty and CLIP score for each language. Combined with English and Finnish, this gives a 25-language evaluation of prompt-level uncertainty under multilingual distribution shift.

\autoref{fig:uncertainty_all_languages} shows a strong negative correlation ($r=-0.79$) between mean EMoE uncertainty and mean CLIP score across languages. Languages with higher uncertainty tend to produce lower-alignment images, while languages with lower uncertainty tend to produce better-aligned images. This provides evidence that EMoE's latent-space epistemic uncertainty estimates capture systematic differences in text-to-image generation reliability across languages.

The size of each point in \autoref{fig:uncertainty_all_languages} corresponds to the number of native speakers for that language. Languages with larger speaker populations generally show lower uncertainty and higher CLIP scores, suggesting a relationship between language prevalence and model coverage. The result shows that EMoE exposes language-dependent reliability differences from a model-internal uncertainty signal. The same negative relationship between uncertainty and generation quality is also observed using Image Reward in \autoref{sec:bias_clip}.

\subsection{Ablation} \label{sec:ablations}
\begin{figure*}[!t]
  \centering

  \begin{minipage}[t]{0.49\textwidth}
    \centering
    \includegraphics[width=\linewidth]{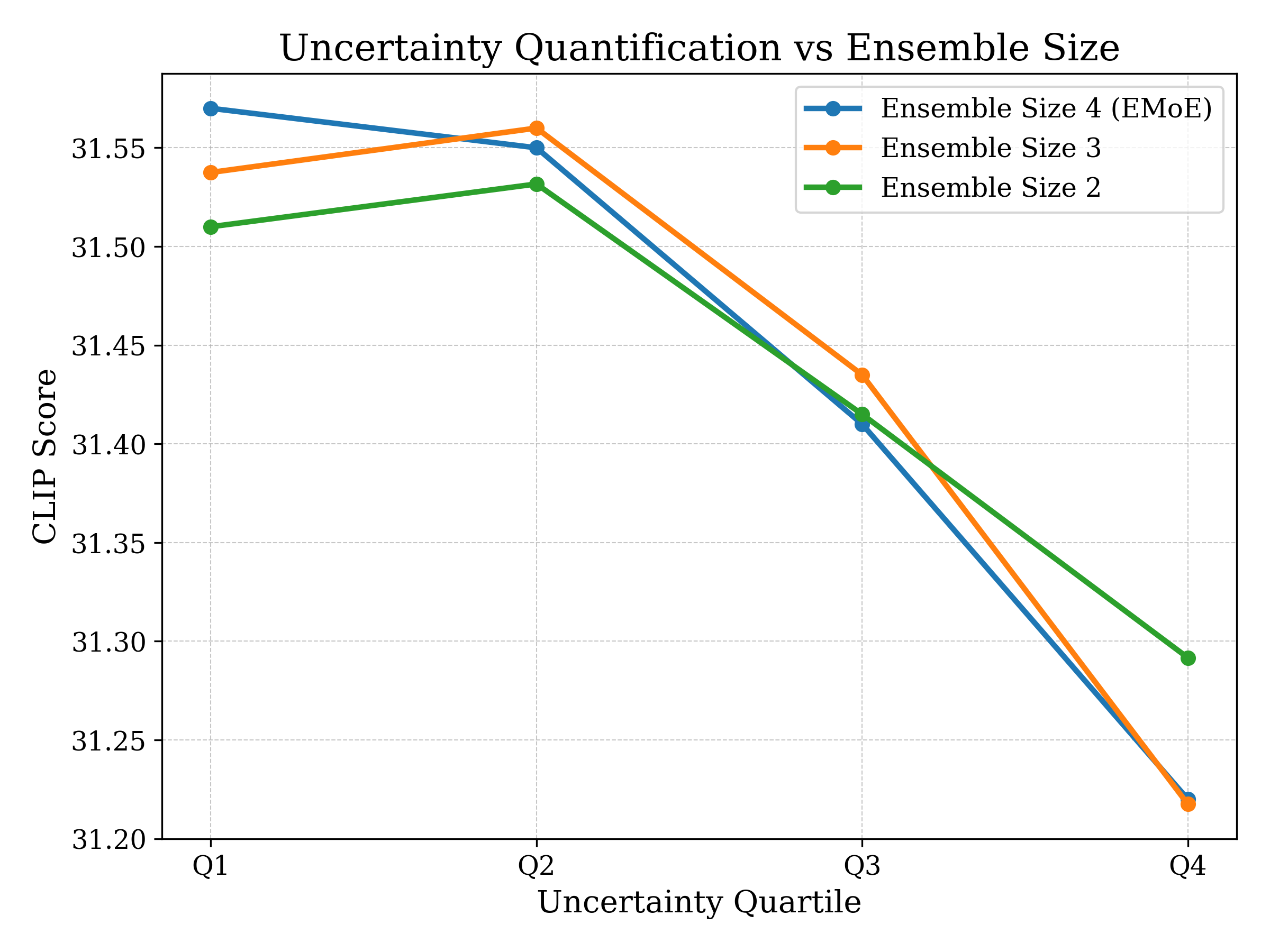}
    \subcaption{Ablation on ensemble size.}
    \label{fig:ablations_ensemble_size}
  \end{minipage}\hfill
  \begin{minipage}[t]{0.49\textwidth}
    \centering
    \includegraphics[width=\linewidth]{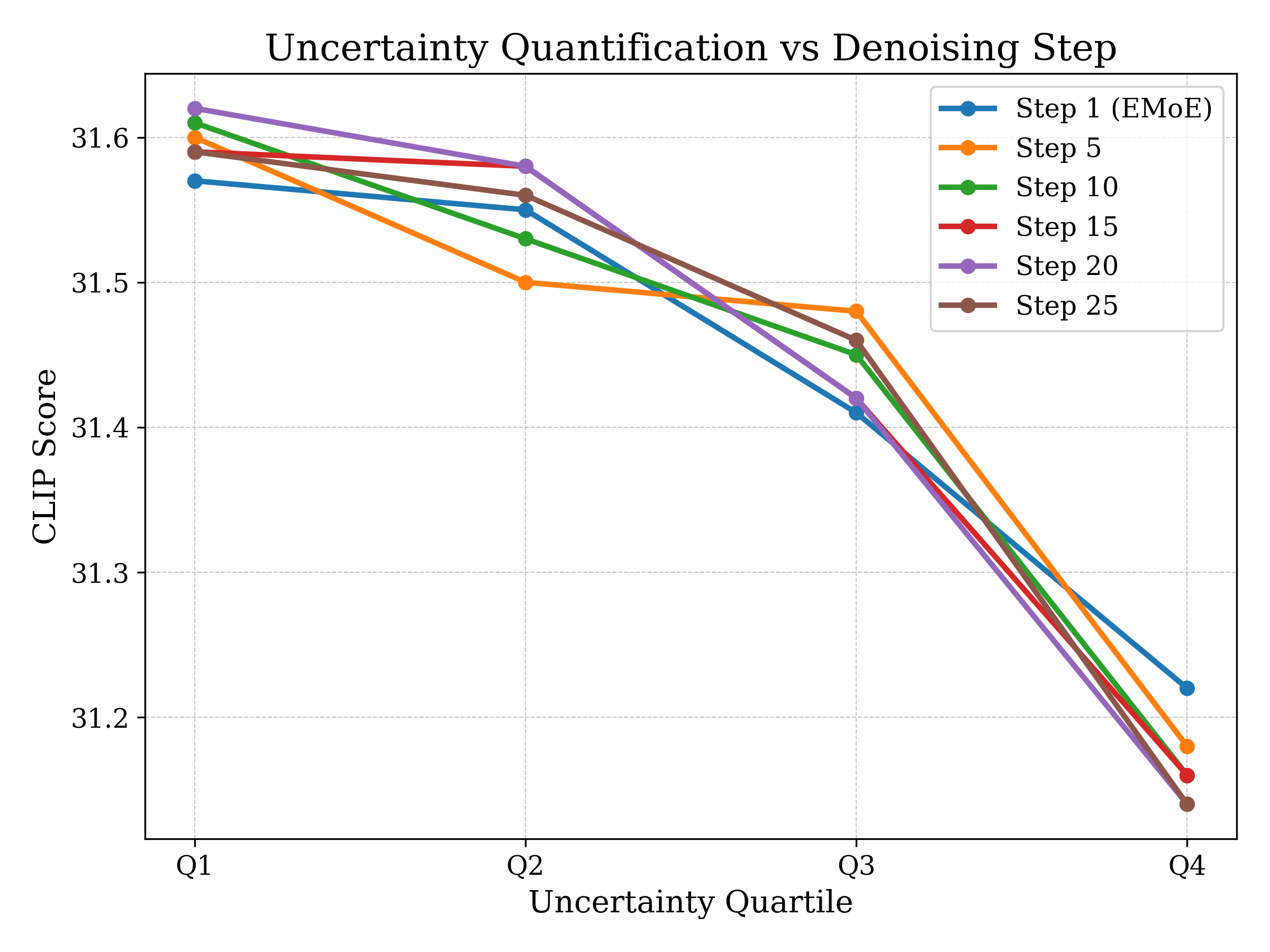}
    \subcaption{Ablation on denoising steps.}
    \label{fig:ablation_denoising_step}
  \end{minipage}

  \vspace{0.1cm}

  \begin{minipage}[t]{0.49\textwidth}
    \centering
    \includegraphics[width=\linewidth]{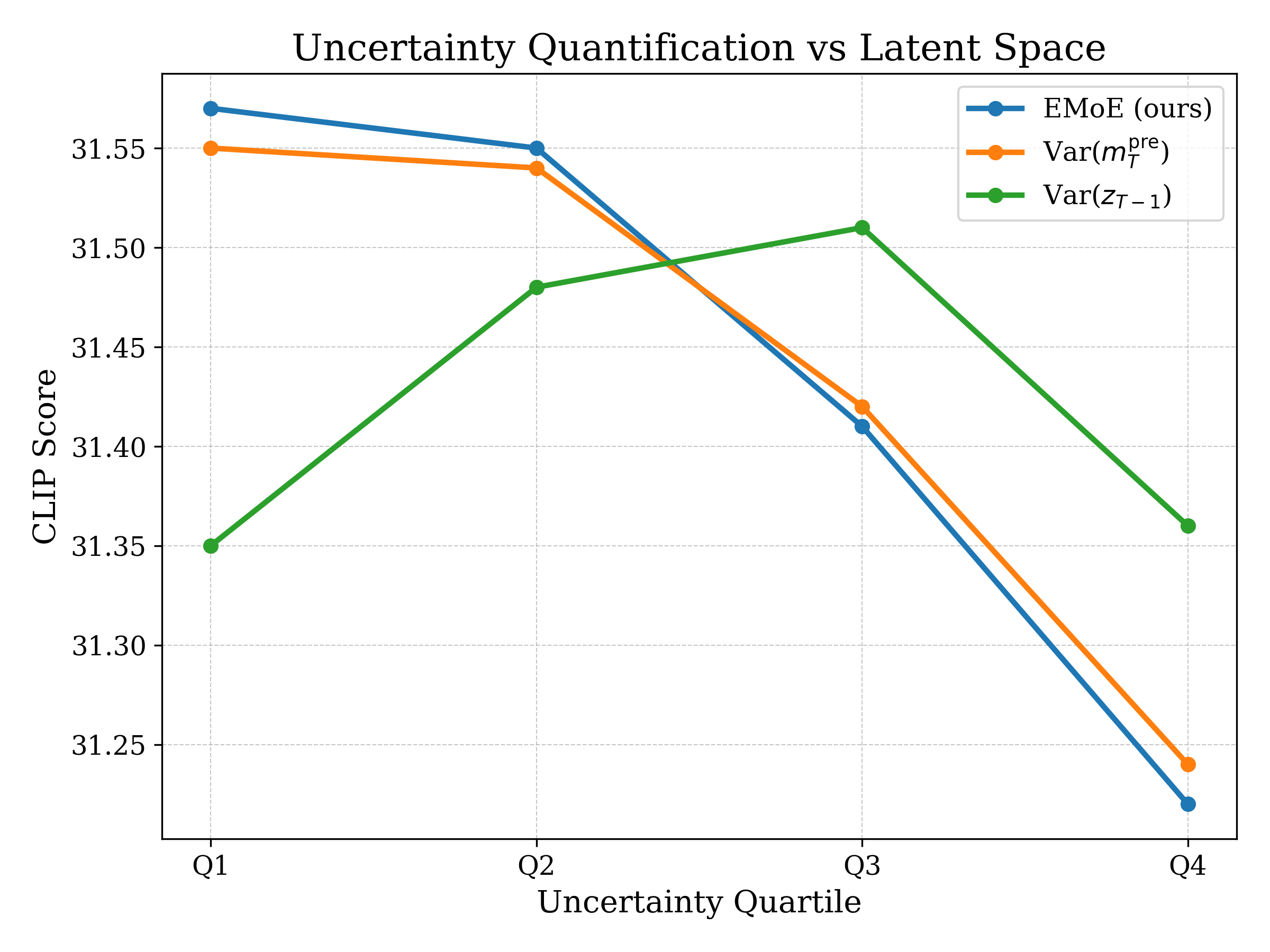}
    \subcaption{Ablation on latent spaces.}
    \label{fig:ablation_other_spaces}
  \end{minipage}\hfill
  \begin{minipage}[t]{0.49\textwidth}
    \centering
    \includegraphics[width=\linewidth]{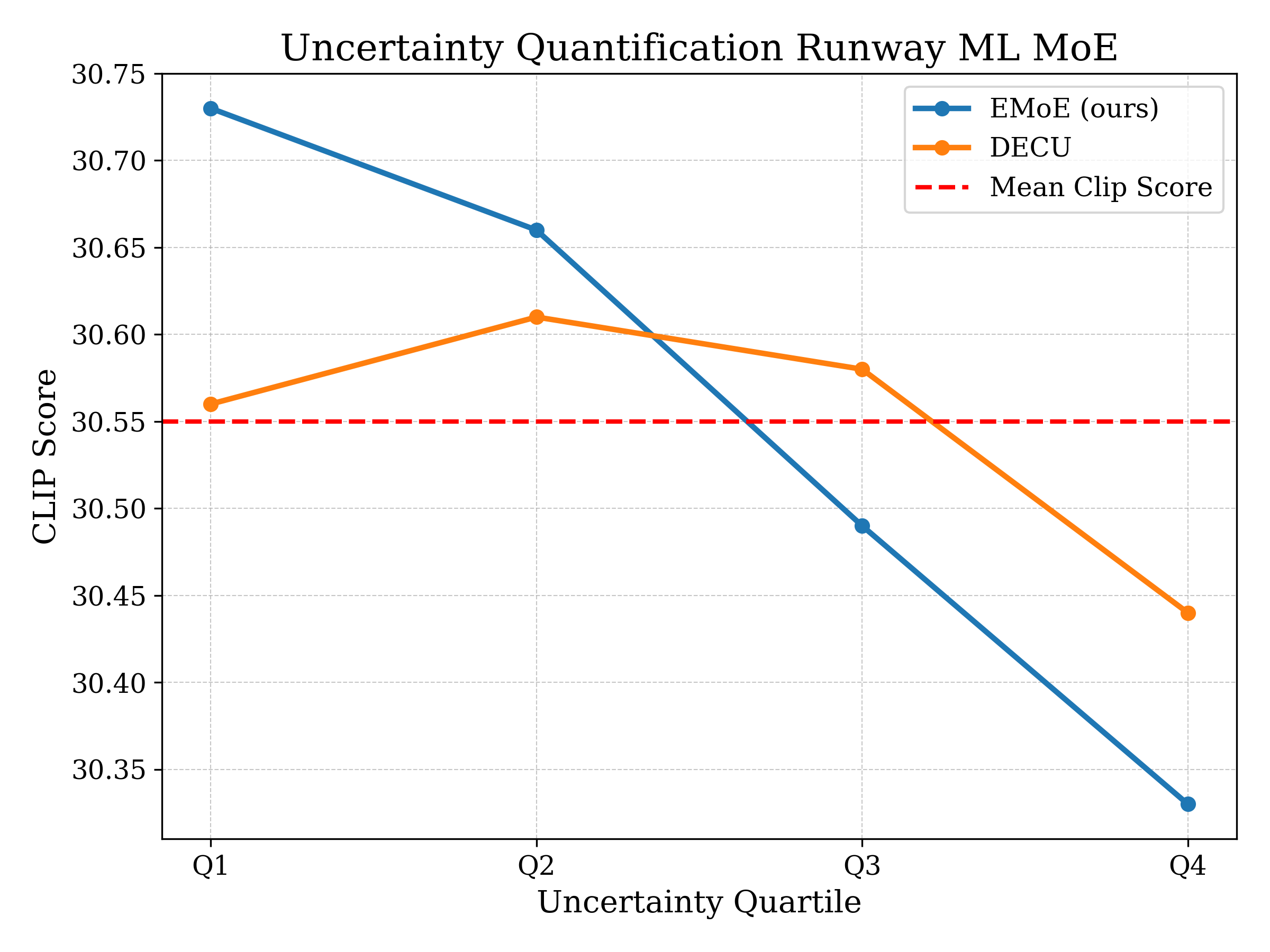}
    \subcaption{Runway ML MoE.}
    \label{fig:ablation_runway_segmoe}
  \end{minipage}

  \caption{Ablation studies validating EMoE's main design choices: ensemble size (a), denoising step (b), latent-space representation (c), and robustness to a different MoE architecture (d).}
  \label{fig:combined_ablations}
\end{figure*}
We conduct four ablations on 40,000 English prompts to evaluate EMoE's main design choices: ensemble size, denoising step, latent-space representation, and model architecture.

First, we vary ensemble size by constructing two- and three-expert ensembles from the four available components and averaging over all permutations (\autoref{fig:ablations_ensemble_size}). The four-expert model gives the most consistent monotonic relationship between uncertainty quartile and CLIP score.

Second, we vary the denoising step used for uncertainty estimation. As shown in \autoref{fig:ablation_denoising_step}, EMoE remains informative across steps, but we select the first step because it provides the earliest opportunity for abstention, or filtering before full generation.

Third, we compare latent representations: $\mathrm{Var}(m_T^{\mathrm{pre}})$, $\mathrm{Var}(m_T^{\mathrm{post}})$, and $\mathrm{Var}(z_{T-1})$. \autoref{fig:ablation_other_spaces} shows that $\mathrm{Var}(z_{T-1})$ is sub-optimal, consistent with DECU \citep{berryshedding}, while $\mathrm{Var}(m_T^{\mathrm{pre}})$ and $\mathrm{Var}(m_T^{\mathrm{post}})$ perform similarly. We use $\mathrm{Var}(m_T^{\mathrm{post}})$ because the mid-block representation has integrated and refined information from the downsampling path.

Finally, \autoref{fig:ablation_runway_segmoe} shows that EMoE remains informative on the Runway ML MoE (\autoref{sec:model_cards}). Since these experts are checkpoints from the same model family, this result suggests that EMoE is not merely measuring extreme stylistic disagreement, but can also capture epistemic variation among closely related expert components.

\section{Related Works} \label{sec:related}
Estimating uncertainty in generative models is difficult because outputs are high-dimensional, multimodal, and costly to sample. Prior work uses Bayesian approximations, ensembles, auxiliary networks, and predictive-distribution-based methods for image and text generation \citep{malinin2020uncertainty, berryshedding, chan2024hyper, liu2024uncertainty}. In diffusion models, DECU estimates epistemic uncertainty from internal representations but requires additional training \citep{berryshedding}, while Hyper-diffusion estimates epistemic and aleatoric uncertainty using hyper-networks \citep{chan2024hyper}. More broadly, epistemic uncertainty is commonly estimated through disagreement among model components, including Bayesian neural networks, Monte Carlo dropout, and deep ensembles \citep{gal2017deep, kendall2017uncertainties, kirsch2019batchbald, lakshminarayanan2017simple}. EMoE extends this disagreement-based view to pre-trained MoE diffusion models by separating expert-specific paths inside a text-conditioned denoiser, without training additional uncertainty modules or sampling multiple complete generations.

MoE architectures improve scalability by routing inputs through specialized components \citep{jacobs1991adaptive, shazeer2017outrageously, fedus2022switch}. Prior uncertainty-aware MoE methods \citep{zheng2019self, luttner2023training, zhang2024efficient} do not address zero-shot epistemic uncertainty in text-to-image diffusion. EMoE connects these directions by using MoE expert structure as a source of latent-space epistemic disagreement.

Text-to-image models often inherit biases from web-scale training data, including uneven coverage across languages, regions, and demographic groups \citep{chinchure2023tibet, alabdulmohsin2024clip}. Multilingual vision-language evaluation often uses translated caption datasets, such as COCO-35L \citep{thapliyal2022crossmodal}. Unlike evaluation-only studies, EMoE provides a model-internal uncertainty signal for diagnosing language-dependent reliability gaps without access to training data.

\section{Conclusions} \label{sec:conclusions}

We introduced EMoE, a training-free framework for estimating epistemic uncertainty in the latent space of text-to-image diffusion models. EMoE separates expert-specific computation paths in pre-trained MoE diffusion models and measures disagreement inside the denoiser. By fixing the initial noise across expert paths, EMoE controls for sampling variation and obtains a prompt-level uncertainty estimate early in generation.

Across COCO and CC3M, lower EMoE uncertainty corresponds to stronger text-image alignment. Under multilingual prompt shifts, EMoE detects elevated uncertainty for underrepresented language inputs and finer lexical effects, such as reduced uncertainty for Finnish prompts containing shared-vocabulary terms with English. These results support EMoE as a practical diagnostic for prompt reliability, model coverage, and language-dependent bias.

\textbf{Limitations.} EMoE requires a pre-trained MoE model or expert checkpoints, additional memory for expert-conditioned paths, and task-relevant expert diversity. Our Runway ML MoE ablation suggests that extreme stylistic diversity is not required, but disagreement will be uninformative if experts behave nearly identically. EMoE estimates uncertainty in the denoiser's latent space and does not provide a full image-space decomposition of aleatoric and epistemic uncertainty. Future work may explore improved expert selection, aleatoric uncertainty estimation, and extensions to other generative modalities.

\bibliographystyle{plainnat}
\bibliography{references}

\newpage
\appendix
\onecolumn
\section{Compute Details} \label{sec:compute_details}
\begin{wraptable}{r}{0.6\textwidth}
\begin{minipage}{0.6\textwidth}

\centering
\caption{Computational requirements.}
\begin{tabular}{ccc}
\hline
\textbf{Dataset} & \textbf{Run Time} & \textbf{Storage}\\ \hline
English 40k Prompts & 200 gpu hrs & 6 TB\\ \hline
Finnish 10k Prompts & 50 gpu hrs & 1.5 TB\\ \hline
Other Languages 1k Prompts & 5 gpu hrs & 150 GB\\ \hline
\end{tabular}
\label{tab:compute_demands}

\vspace{1.0cm}
\centering
\caption{Generation times for baseline (SegMoE) and two variants of EMoE. Reported times are $\mu\pm\sigma$.}
\begin{tabular}{cc}
\hline
\textbf{Model} & \textbf{Generation Time} \\ \hline
SegMoE & 3.58 $\pm$ 0.54 secs \\ \hline
EMoE & 12.32 $\pm$ 4.6 secs \\ \hline
Fast EMoE & 4.5 $\pm$ 0.16 secs\\ \hline
\end{tabular}
\label{tab:compute_times}
\end{minipage}
\end{wraptable}
We used the same set of hyperparameters as in the Stable Diffusion model described by \citet{segmoe}. All experiments use DDIM sampling with 25 denoising steps and the standard $\epsilon$-prediction parameterization, where the model predicts the noise added to the latent state at each timestep, following the formulation in \citet{rombach2022high}. Minor changes were made to both the \href{https://github.com/segmind/segmoe}{SegMoE} and \href{https://huggingface.co/}{Diffusers} codebases to disentangle the MoE, with specific modifications to incorporate EMoE. Our infrastructure included an AMD Milan 7413 CPU running at 2.65 GHz, with a 128M L3 cache, and an NVIDIA A100 GPU with 40 GB of memory. The wall clock time required to collect each dataset and the memory usage are provided in  \autoref{tab:compute_demands}. The parameter count for the SegMoE model is 1.63 billion parameters, while a single model contains 1.07 billion parameters. This highlights the efficiency of using a sparse MoE approach compared to creating 4 distinct models, as the SegMoE model is only 153\% the size of a single model, rather than 400\%. When running the SegMoE model in its standard mode, generating an image from one prompt takes an average of 3.58 seconds. In comparison, using EMoE typically requires an average of 12.32 seconds to generate four images from a single prompt. However, when only a single image per prompt is required, EMoE can be optimized by estimating epistemic uncertainty during the initial diffusion step. Once uncertainty is computed, standard MoE-based image generation proceeds after pruning unnecessary computational paths. This optimized version, Fast EMoE, achieves an average generation time of 4.5 seconds. This is substantially more efficient than loading four independent diffusion models and sampling from each separately, which would require significantly greater memory and compute resources unless extensive parallel hardware is available. \autoref{tab:compute_times} provides further details. Note that uncertainty reported across all experiments is calculated as $\sqrt{d_{midsize}}\times\text{EU}(y)$, where $d_{midsize}=1280\times8\times8$.

\section{Bias in CLIP score} \label{sec:bias_clip}

CLIP score, despite its known biases \citep{chinchure2023tibet, alabdulmohsin2024clip}, remains a widely used metric for evaluating alignment between text prompts and generated images, alongside FID \citep{shi2020improving, kumari2023multi}. Both metrics rely on auxiliary models (CLIP and Inception, respectively), making them susceptible to inherited biases. FID also requires many samples for reliable estimation and does not measure text-image alignment, which is central to our setting. In contrast, CLIP score provides a direct measure of prompt-image alignment with fewer samples \citep{kawar2023imagic, ho2022imagen}. We therefore prioritize CLIP score due to its relevance to our objective and its broad use in text-to-image evaluation.

\begin{wraptable}{r}{0.52\textwidth}
\vskip -0.2in
\centering
\caption{Comparison of Aesthetic Score and Image Reward with mean uncertainty $\pm$ standard deviation between Finnish and English prompts. Image Reward aligns with the lower reliability and higher uncertainty of Finnish prompts, while Aesthetic Score evaluates image-only visual appeal.}
\resizebox{0.52\textwidth}{!}{
\begin{tabular}{cccc}
\hline
\textbf{Language} & \textbf{Aesthetic Score \textuparrow} & \textbf{Image Reward \textuparrow} & \textbf{Uncertainty}\\ \hline
Finnish & 5.917 & -2.143 & 0.48 $\pm$ 0.19  \\ \hline
English & 5.738 & 0.270 & 0.37 $\pm$ 0.14 \\ \hline
\end{tabular}
}
\label{tab:english_vs_finnish_not_clip}
\vskip -0.2in
\end{wraptable}

To address metric bias concerns, we also evaluate generated images using the Aesthetic Score Predictor and Image Reward \citep{schuhmann2022laion, xu2024imagereward}. Aesthetic Score estimates image-only visual appeal on a scale from 1 to 10, while Image Reward encodes human preferences conditioned on both the image and prompt. Thus, Image Reward is more directly aligned with our goal of evaluating prompt-conditioned generation quality, whereas Aesthetic Score can be high even when the generated image is poorly aligned with the prompt.

\begin{algorithm}[ht] 
\caption{Epistemic Mixture of Experts (EMoE)} 
\label{alg:emoe} 
\begin{algorithmic}[1] 
\STATE \textbf{Input:} Initial noise $z_T \sim \mathcal{N}(\mathbf{0}, \mathbf{I})$, total steps $T$, pre-trained experts $E = \{e_1, e_2, \dots, e_M\}$, prompt $y$ 
\FOR{$t = T$ to $1$} 
\IF{$t = T$} 
\STATE \textbf{Separate Experts:} 
\FOR{each expert $e_i \in E$} 
\STATE Pass $z_T$ and prompt $y$ through $e_i$'s first cross-attention layer $CA^i$ to arrive at $M$ distinct latent representations. 
\STATE Subsequent sparse MoE layers are processed as \autoref{eq:expert_weighted_sum}. 
\STATE Extract the $\mathrm{mid}$-block latent representation for each expert $m^{post,i}_T$. 
\ENDFOR 
\STATE Compute epistemic uncertainty $\text{EU}(y)$ as defined in \autoref{eq:emoe_estimator}. 
\STATE Output $M$ different $\mathbf{z}_{t-1}^{i}$, one for each expert. 
\ELSE 
\STATE \textbf{Mixture of Experts Rollout:} 
\FOR{$i\in\{1,...,M\}$} 
\STATE Update latent variable for each expert: \[ \mathbf{z}_{t-1}^{i} \sim p_{\theta}(\mathbf{z}^i_{t-1} | \mathbf{z}^i_t, y) \] 
\STATE Pass $\mathbf{z}^i_{t-1}$ and $y$ through our reverse diffusion process, as a standard MoE (e.g. experts are aggregated as \autoref{eq:expert_weighted_sum}). This is shown in \autoref{fig:flowchart} in $\cblock{spec8}$ and $\cblock{spec1}$. 
\ENDFOR 
\ENDIF 
\ENDFOR 
\STATE \textbf{Output:} $M$ reconstructed latent variables $\mathbf{z}^i_0$ and $\text{EU}(y)$. 
\end{algorithmic} 
\end{algorithm}

We repeat the Finnish-English comparison using these metrics. Image Reward supports the same conclusion as CLIP: Finnish prompts receive lower scores and higher uncertainty than English prompts (\autoref{tab:english_vs_finnish_not_clip}). In contrast, Aesthetic Score is higher for Finnish prompts. This does not contradict our interpretation because Aesthetic Score ignores the prompt and measures only visual appeal. Prior work has shown that text-to-image models can produce visually pleasing default images even from weak, underspecified, or uninformative prompts \citep{simonen2026exploration}. Thus, an image may be aesthetically appealing while still failing to match the intended prompt, which explains why Aesthetic Score can diverge from CLIP and Image Reward in our setting.

\begin{wrapfigure}{r}{0.45\textwidth}
    \vskip -0.2in
    \centering
    \includegraphics[width=0.45\textwidth]{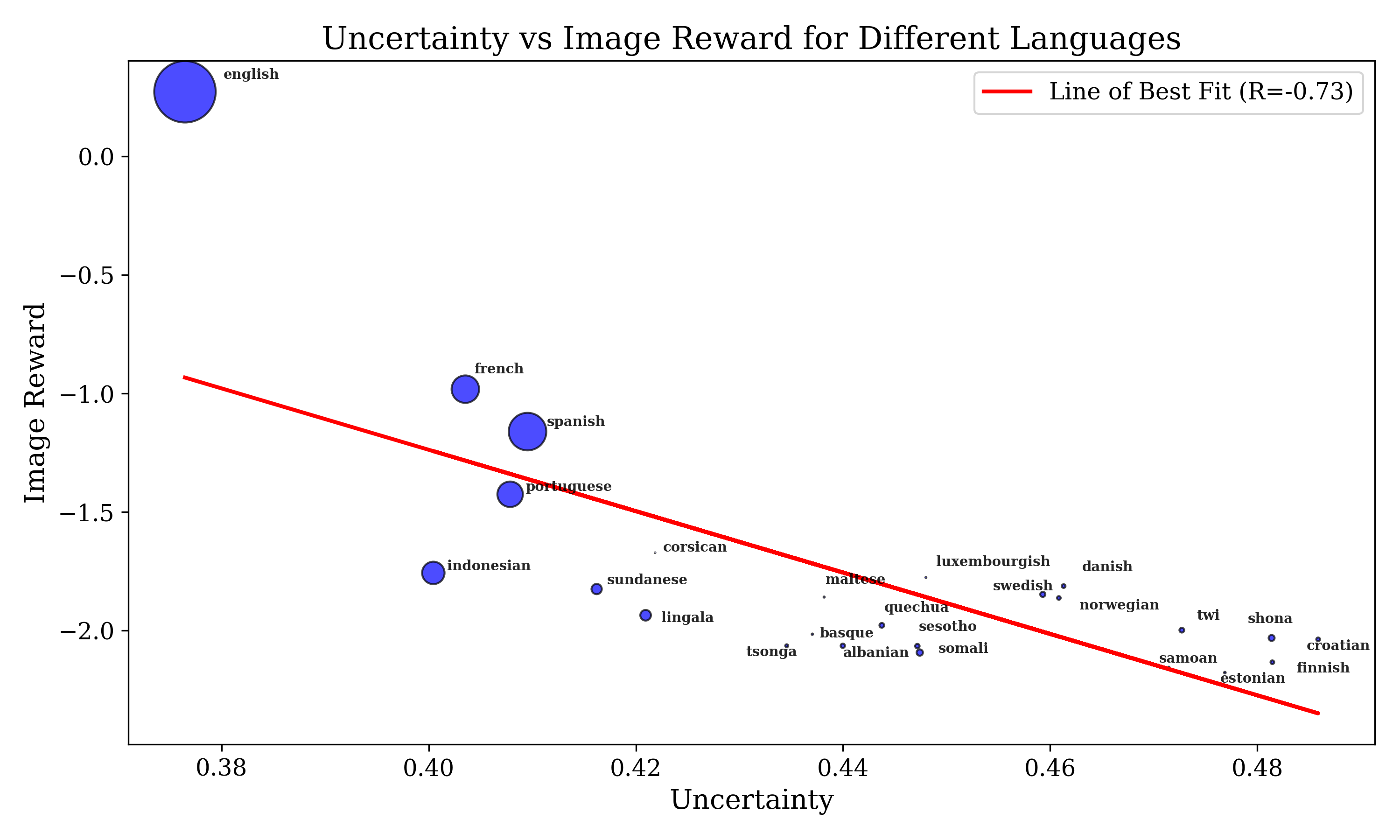}
    \caption{Negative correlation between uncertainty and image quality across prompts translated into 25 languages. EMoE demonstrates a strong negative correlation ($r=-0.73$) between uncertainty and Image Reward.}
    \label{fig:all_languages_reward}
    \vskip -0.4in
\end{wrapfigure}

Finally, we evaluate the correlation between EMoE uncertainty and Image Reward across all languages (\autoref{fig:all_languages_reward}). Consistent with the CLIP-based analysis, higher uncertainty is associated with lower Image Reward ($r=-0.73$). This confirms that the multilingual uncertainty trend is not specific to CLIP and further supports EMoE's ability to detect language-dependent reliability gaps in text-to-image generation.

\section{CC3M Dataset} \label{sec:additional_dataset}

\begin{figure}[t]
\centering

\begin{minipage}[t]{0.48\textwidth}
\vskip 0.2in
    \centering
    \includegraphics[width=\linewidth]{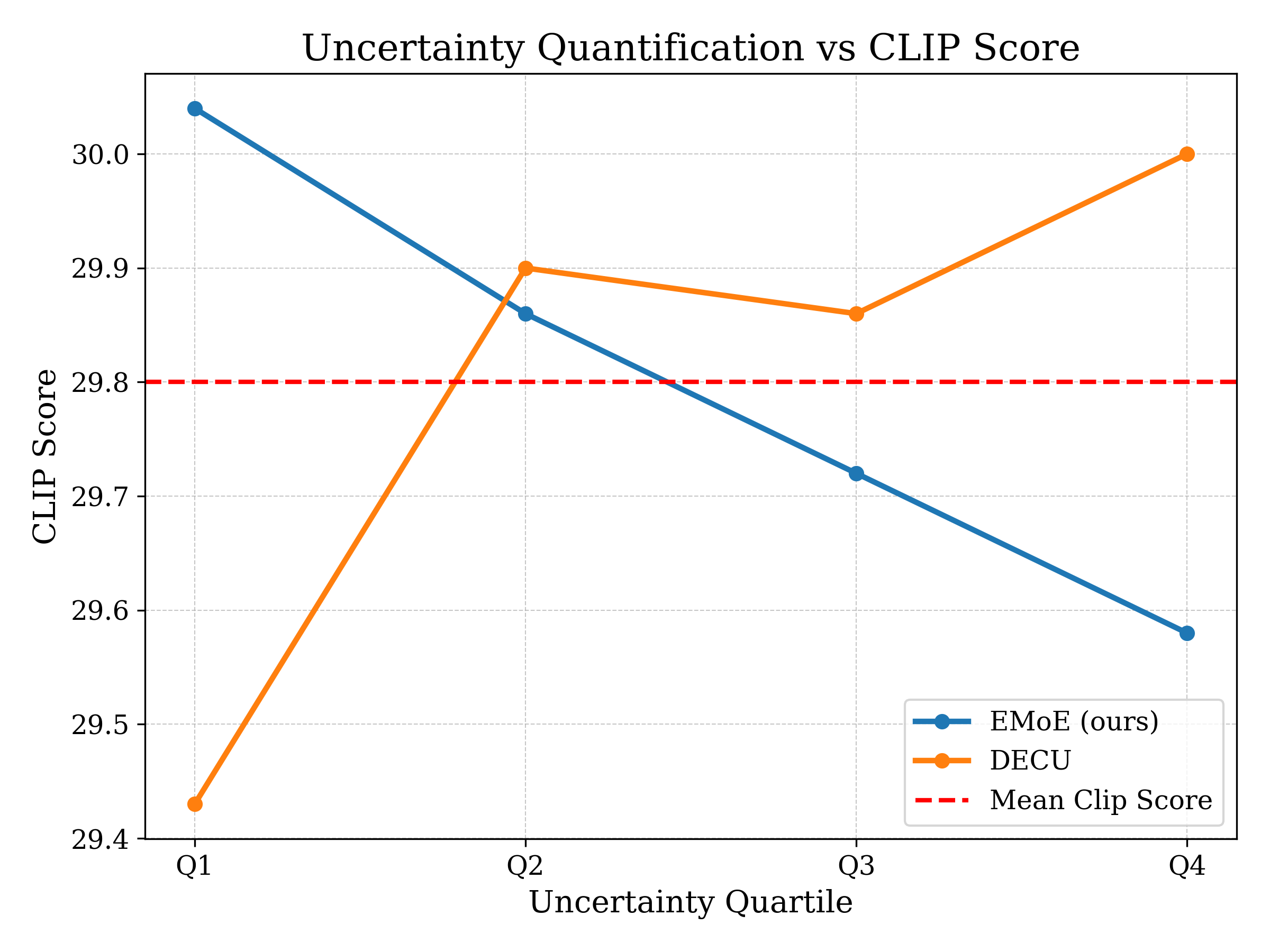}
    \caption{CLIP score on the CC3M dataset across different uncertainty quartiles. EMoE accurately attributes prompts that produce images with high CLIP scores to low uncertainty, unlike DECU. The red line indicates the average CLIP score across all quartiles.}
    \label{fig:cc3m_clip}
    \vskip 0.2in
\end{minipage}%
\hfill
\begin{minipage}[t]{0.48\textwidth}
\centering
\vskip 0.2in %

\captionsetup[table]{name=Table}
\captionof{table}{CLIP score, Aesthetic score \& Image Reward on each uncertainty quartile, using EMoE, on the English 10k prompt dataset for CC3M dataset.}
\resizebox{\linewidth}{!}{
\begin{tabular}{cccc}
\hline
\textbf{Quartile}& \textbf{CLIP Score \textuparrow} & \textbf{Aesthetic Score \textuparrow} & \textbf{Image Reward \textuparrow}  \\ \hline
Q1 & 30.07 & 5.75 & -0.05\\ \hline
Q2 & 29.83 & 5.63 & -0.12\\ \hline
Q3 & 29.72 & 5.60 & -0.17\\ \hline
Q4 & 29.57 & 5.44 & -0.26\\ \hline
\end{tabular}
}
\label{tab:more_scores_cc3m} %
\end{minipage}

\end{figure}

To further validate our results, we extended the analysis of EMoE to the CC3M dataset \citep{sharma2018conceptual}. The results are presented in \autoref{fig:cc3m_clip} and \hyperref[tab:more_scores_cc3m]{\Cref{tab:more_scores_cc3m}}. We randomly sampled 10,000 prompts from the dataset and repeated our analysis on English prompts. Our findings demonstrate that EMoE continues to perform robustly on the CC3M dataset. Specifically, prompts with lower uncertainty produced higher-quality images compared to those with higher uncertainty. These results not only reaffirm the robustness of our findings on the COCO dataset but also provide additional validation across a different dataset. Importantly, they suggest that EMoE's ability to capture reliability variation is consistent and not merely an artifact of the COCO dataset. This further strengthens the argument for EMoE’s effectiveness in diverse settings and its capacity for detecting biases in real-world data.

\section{Intuition and Identities for the EMoE Estimator}
\label{sec:gp_analogy}

This section provides intuition for the EMoE estimator through the lens of ensembles and Gaussian Processes (GPs). It should not be read as a proof that pre-trained MoE experts are exact posterior samples. Rather, it motivates the use of expert disagreement as an epistemic signal.

In an ideal Bayesian ensemble, different predictors can be viewed as samples from a posterior over functions. For a fixed input $y$, the variance of their predictions estimates posterior uncertainty over the function value at $y$. In a GP, this uncertainty corresponds to the marginal posterior variance of the latent function. Thus, disagreement among plausible predictors can be interpreted as epistemic uncertainty, as in ensemble-based methods \citep{gal2017deep, depeweg2018decomposition, berry2023normalizing, lakshminarayanan2017simple}.

EMoE differs from this idealized Bayesian setting in two important ways. First, its experts come from pre-trained MoE checkpoints rather than from i.i.d. posterior samples. Second, EMoE measures uncertainty in the denoiser's latent space, not directly in image space, by separating expert-conditioned paths for the same prompt $y$ and initial noise $z_T$. Therefore, the Bayesian ensemble argument below is used only as motivation. The trace-covariance identity that follows holds directly for the EMoE estimator.

For intuition, let $\mathcal{F} = \{ f_{\theta_i} \}_{i=1}^{N}$ denote an ensemble of $N$ predictors, where each model $f_{\theta_i}: \mathcal{Y} \to \mathbb{R}$ is parameterized by $\theta_i$. In an ideal Bayesian setting, each $\theta_i$ can be viewed as a sample from a posterior over parameters. The ensemble prediction is
\[
\hat{f}(y) = \frac{1}{N} \sum_{i=1}^{N} f_{\theta_i}(y),
\]
where $y \in \mathcal{Y}$ is an input.

A \textbf{Gaussian Process} is a collection of random variables such that any finite subset follows a joint Gaussian distribution. Conditional on data, a GP posterior can be written as
\[
f(y) \mid \mathcal{D} \sim \mathcal{GP}(\mu_{\mathcal{D}}(y), k_{\mathcal{D}}(y,y')),
\]
where $\mu_{\mathcal{D}}(y)$ is the posterior mean function and $k_{\mathcal{D}}(y,y')$ is the posterior covariance function:
\[
\mu_{\mathcal{D}}(y) = \mathbb{E}[f(y)\mid \mathcal{D}], 
\qquad 
k_{\mathcal{D}}(y,y') =
\mathbb{E}\left[
(f(y)-\mu_{\mathcal{D}}(y))
(f(y')-\mu_{\mathcal{D}}(y'))
\mid \mathcal{D}
\right].
\]
The marginal posterior variance $k_{\mathcal{D}}(y,y)$ therefore measures uncertainty over the latent function value at input $y$.

\textbf{Proposition 1.}
Suppose that, for a fixed input $y$, the expert predictions
\[
f_{\theta_1}(y), \ldots, f_{\theta_N}(y)
\]
are i.i.d. samples from a posterior distribution over function values with finite variance. Then the empirical ensemble variance
\[
\widehat{\mathrm{Var}}[f(y)] 
= \frac{1}{N} \sum_{i=1}^{N} 
\left(f_{\theta_i}(y) - \hat{f}(y)\right)^2
\]
is a consistent estimator of the posterior variance over function values at $y$. In the GP case, if $k_{\mathcal{D}}$ denotes the posterior covariance function, this limiting quantity is the marginal posterior variance $k_{\mathcal{D}}(y,y)$.

\textbf{Proof.}
Let
\[
\mu(y) = \mathbb{E}_{\theta}[f_{\theta}(y)],
\qquad
\sigma^2(y) = \mathrm{Var}_{\theta}[f_{\theta}(y)].
\]
By the law of large numbers,
\[
\hat{f}(y) = \frac{1}{N}\sum_{i=1}^{N} f_{\theta_i}(y)
\longrightarrow \mu(y)
\]
almost surely as $N \to \infty$. Similarly,
\[
\frac{1}{N}\sum_{i=1}^{N} f_{\theta_i}(y)^2
\longrightarrow 
\mathbb{E}_{\theta}[f_{\theta}(y)^2].
\]
Therefore,
\[
\widehat{\mathrm{Var}}[f(y)]
=
\frac{1}{N}\sum_{i=1}^{N} f_{\theta_i}(y)^2 - \hat{f}(y)^2
\longrightarrow
\mathbb{E}_{\theta}[f_{\theta}(y)^2] - \mu(y)^2
=
\sigma^2(y).
\]
If the distribution over function values is induced by a GP posterior, then $\sigma^2(y)=k_{\mathcal{D}}(y,y)$. Thus, under the idealized assumption that ensemble members are posterior samples, disagreement among ensemble predictions consistently estimates epistemic uncertainty over function values.

In EMoE, the experts are not assumed to satisfy this posterior-sampling condition. Instead, this result motivates the use of expert disagreement as an epistemic signal. EMoE applies this idea to expert-conditioned denoising paths in latent space, measuring disagreement between representations produced from the same prompt and initial noise.

\textbf{EMoE as latent trace covariance.}
We now state an identity that holds directly for the EMoE estimator. Let
\[
m_i = m^{\mathrm{post}}_{T,i}(y,z_T) \in \mathbb{R}^{D_{\mathrm{mid}}}
\]
denote the mid-block representation produced by expert path $i$ for the same prompt $y$ and initial noise $z_T$, and let
\[
\bar{m} = \frac{1}{M}\sum_{i=1}^{M}m_i.
\]
Using population variance over experts, EMoE satisfies
\[
\mathrm{EU}(y)
=
\frac{1}{D_{\mathrm{mid}}}
\sum_{d=1}^{D_{\mathrm{mid}}}
\mathrm{Var}_{i=1,\ldots,M}
\left(m_{i,d}\right)
=
\frac{1}{D_{\mathrm{mid}}M}
\sum_{i=1}^{M}
\|m_i-\bar{m}\|_2^2.
\]
Equivalently,
\[
\mathrm{EU}(y)
=
\frac{1}{2D_{\mathrm{mid}}M^2}
\sum_{i=1}^{M}
\sum_{j=1}^{M}
\|m_i-m_j\|_2^2.
\]
Thus, EMoE can be interpreted as the normalized trace covariance of expert-conditioned latent representations, or equivalently as the average pairwise squared disagreement between expert paths.

\textbf{Proof.}
The first equality follows by summing coordinate-wise variance over latent dimensions:
\[
\sum_{d=1}^{D_{\mathrm{mid}}}
\mathrm{Var}_{i}(m_{i,d})
=
\frac{1}{M}\sum_{i=1}^{M}\|m_i-\bar{m}\|_2^2.
\]
The pairwise form follows from the standard identity
\[
\sum_{i=1}^{M}\|m_i-\bar{m}\|_2^2
=
\frac{1}{2M}
\sum_{i=1}^{M}
\sum_{j=1}^{M}
\|m_i-m_j\|_2^2.
\]
Substituting this identity gives the result.

\textbf{Why fixing the initial noise matters.}
Let $Z$ denote the initial diffusion noise, let $I$ denote a uniformly sampled expert path, and let $\mathrm{Tr}$ denote the trace operator. For a fixed prompt $y$, the law of total variance gives
\[
\mathrm{Tr}\,\mathrm{Cov}_{I,Z}
\left[m_I(y,Z)\right]
=
\mathbb{E}_{Z}
\left[
\mathrm{Tr}\,\mathrm{Cov}_{I}
\left[m_I(y,Z)\mid Z\right]
\right]
+
\mathrm{Tr}\,\mathrm{Cov}_{Z}
\left[
\mathbb{E}_{I}
\left[m_I(y,Z)\mid Z\right]
\right].
\]
The first term is expert disagreement at fixed noise, averaged over possible initial noise values. The second term is variation caused by changing the initial noise. This decomposition clarifies why EMoE holds the initial noise fixed across expert paths: otherwise, the measured variance could mix expert disagreement with stochastic variation from different diffusion trajectories.

In our experiments, EMoE samples one initial noise $z_T$ per prompt and holds it fixed across expert paths. Thus, for the chosen noise seed, EMoE measures the conditional expert-disagreement component
\[
\mathrm{EU}(y;z_T)
=
\frac{1}{D_{\mathrm{mid}}}
\mathrm{Tr}\,\mathrm{Cov}_{I}
\left[
m_I(y,Z)\mid Z=z_T
\right].
\]
Therefore, the variance measured by EMoE is induced by expert-specific latent representations rather than by different stochastic generations.

\newpage
\section{Gates Without Training} \label{sec:gates_train}
\begin{wrapfigure}{r}{0.45\textwidth}
    \vskip -0.4in
    \centering
    \resizebox{0.45\textwidth}{!}{\begin{tikzpicture}[
  every node/.style={font=\small, align=center},
  box/.style={rectangle, draw, minimum width=2.5cm, minimum height=1cm, rounded corners, text centered, text width=2.5cm},
  arrow/.style={-latex, thick},
  vector/.style={rectangle, draw, minimum width=0.4cm, minimum height=0.4cm, fill=blue!20}
]

\node[rectangle, draw=none] (descriptor) at (0, 1) {\(\psi^i\)};

\node[rectangle, rounded corners, text centered, text width=1.5cm, minimum height=1cm, draw=\trapezoidcolor!100, fill=\trapezoidcolor!20] (textModel) at (2, 1) {\(\tau_{\theta}\)};
\draw[arrow] (descriptor.east) -- (textModel.west);

\node[rectangle, draw=none] (promptInput) at (2, 2.5) {\(y^j \)};
\draw[arrow] (promptInput.south) -- (textModel.north);

\node[rectangle, draw=none] (gateVector) at (4.8, 1.5) {\(v^i\)};
\foreach \i in {0,1,...,4} {
  \node[vector] (gate\i) at (4 + 0.4*\i, 1) {};
}
\draw[arrow] (textModel.east) -- (gate0.west);

\foreach \i in {0,1,...,4} {
  \node[vector] (rep\i) at (2, -0.5 - 0.4*\i) {};
}
\draw[arrow] (textModel.south) -- (rep0.north);

\node[rectangle, draw=none] (weightAssign) at (4.8, -1.3) {\(w^j_i\)};
\draw[arrow] (gate2.south) -- (weightAssign.north);
\draw[arrow] (rep2.east) -- (weightAssign.west);

\end{tikzpicture}}
    \caption{This figure depicts how to obtain accurate gates without training.}
    \label{fig:gating}
    \vskip -0.2in
\end{wrapfigure}

Each expert is associated with a positive and a negative descriptor, \( \psi^i=\left(p^i,n^i\right) \), which represent what the expert excels at and struggles with modeling, respectively. These descriptors are processed through a pre-trained text model, \( \tau_{\theta} \), to create \emph{gate vectors}, $v^i = [\tau_{\theta}(p^i);\tau_{\theta}(n^i)]$. When a new positive and negative prompt, \( y_j = \left(\phi_j, \nu_j\right) \), is provided to generate an image, the latent representation of these prompts, $l_j=[\tau_{\theta}(\phi_j);\tau_{\theta}(\nu_j)]$ are compared against \( v^i \) and assigned a weight, \( w^j_i \) based on the dot product and a softmax. This process is illustrated in \autoref{fig:gating} and described in \citet{goddard2024arcee}.

\section{Additional Results}
\label{sec:additional}
\begin{table*}[t]
\centering

\begin{minipage}{0.48\textwidth}
\centering
\caption{Mean length of English prompts by uncertainty quartile, $\pm$ standard deviation.}
\label{tab:mean_length_english_prompts}
\resizebox{\linewidth}{!}{%
\begin{tabular}{ccc}
\hline
\textbf{Quartile} & \textbf{Character Count} & \textbf{Word Count} \\ \hline
Q1 & 53.14 $\pm$ 13.50 & 10.58 $\pm$ 2.56 \\ \hline
Q2 & 52.38 $\pm$ 12.94 & 10.47 $\pm$ 2.42 \\ \hline
Q3 & 52.20 $\pm$ 12.81 & 10.43 $\pm$ 2.39 \\ \hline
Q4 & 51.93 $\pm$ 12.32 & 10.34 $\pm$ 2.33 \\ \hline
\end{tabular}%
}
\end{minipage}
\hfill
\begin{minipage}{0.48\textwidth}
\centering
\caption{Mean length of Finnish prompts by uncertainty quartile, $\pm$ standard deviation.}
\label{tab:mean_length_finnish_prompts}
\resizebox{\linewidth}{!}{%
\begin{tabular}{ccc}
\hline
\textbf{Quartile} & \textbf{Character Count} & \textbf{Word Count} \\ \hline
Q1 & 54.94 $\pm$ 17.04 & 6.59 $\pm$ 2.16 \\ \hline
Q2 & 51.26 $\pm$ 14.40 & 6.14 $\pm$ 1.79 \\ \hline
Q3 & 49.67 $\pm$ 14.23 & 5.95 $\pm$ 1.75 \\ \hline
Q4 & 47.97 $\pm$ 13.86 & 5.77 $\pm$ 1.73 \\ \hline
\end{tabular}%
}
\end{minipage}

\end{table*}

\hyperref[tab:mean_length_english_prompts]{\Cref{tab:mean_length_english_prompts}} and 
\hyperref[tab:mean_length_finnish_prompts]{\Cref{tab:mean_length_finnish_prompts}} report prompt-length statistics across EMoE uncertainty quartiles for English and Finnish prompts. In both languages, lower-uncertainty prompts tend to be longer on average. For English, the trend is modest but monotonic, with character count decreasing from 53.14 in Q1 to 51.93 in Q4 and word count decreasing from 10.58 to 10.34. For Finnish, the trend is stronger, with character count decreasing from 54.94 in Q1 to 47.97 in Q4 and word count decreasing from 6.59 to 5.77. The ordered-trend tests in \autoref{sec:stat_analysis} provide statistical support for these patterns.

In \autoref{sec:finnish_prompts}, we report the EMoE Finnish-vs-English uncertainty distribution in the main paper and compare EMoE, DECU, and Gating Entropy quantitatively in \hyperref[tab:finnish_ood_auroc]{\Cref{tab:finnish_ood_auroc}}. Here, we provide the corresponding distribution plots for all three uncertainty measures.

\begin{figure}[t]
    \centering
    \includegraphics[width=\linewidth]{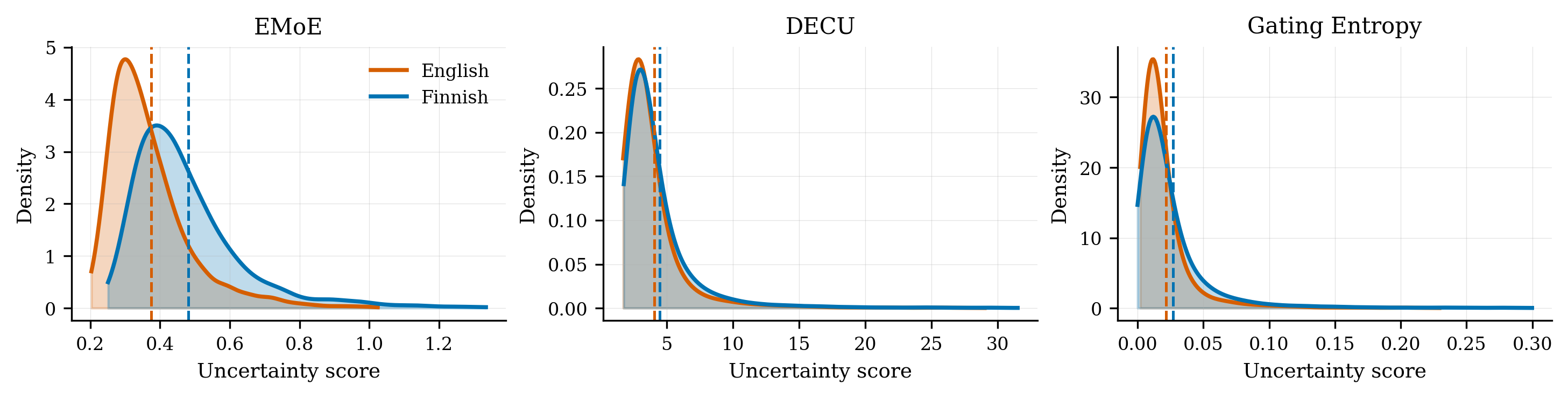}
    \caption{Finnish-vs-English uncertainty distributions for EMoE, DECU, and Gating Entropy. EMoE shows the clearest shift toward higher uncertainty for Finnish prompts, while DECU and Gating Entropy provide weaker separation.}
    \label{fig:finnish_ood_baseline_distributions}
\end{figure}

\autoref{fig:finnish_ood_baseline_distributions} visualizes the same pattern quantified in \hyperref[tab:finnish_ood_auroc]{\Cref{tab:finnish_ood_auroc}}: EMoE separates Finnish and English prompts most clearly, while DECU and Gating Entropy provide weaker separation.

\section{Model Cards} \label{sec:model_cards}

Below are the model parameters for the base SegMoE MoE used in the experiments. We increased the number of experts from 2 to 4 to incorporate more ensemble components. Generally, having a low number of ensemble components (2-10) is sufficient in deep learning to capture model disagreement \citep{osband2016deep, chua2018deep, fujimoto2018addressing}. In addition to the SegMoE base MoE, we also tested EMoE on another MoE model, referred to as Runway ML, where each expert component is a Runway model. The corresponding model card can be found below. This experiment demonstrates the robustness of EMoE across different MoE architectures, showing that EMoE is effective even when components are trained on similar data with similarly initialized weights, as each Runway ML component was fine-tuned on new data from similar initial conditions. 

\newpage
\begin{lstlisting}[style=graybox]
SegMoE MoE

base_model: SG161222/Realistic_Vision_V6.0_B1_noVAE
num_experts: 4
moe_layers: all
num_experts_per_tok: 2
type: sd
experts:
  - source_model: SG161222/Realistic_Vision_V6.0_B1_noVAE
    positive_prompt: "cinematic, portrait, photograph, instagram, fashion, movie, macro shot, 8K, RAW, hyperrealistic, ultra realistic,"
    negative_prompt: " (deformed iris, deformed pupils, semi-realistic, cgi, 3d, render, sketch, cartoon, drawing, anime), text, cropped, out of frame, worst quality, low quality, jpeg artifacts, ugly, duplicate, morbid, mutilated, extra fingers, mutated hands, poorly drawn hands, poorly drawn face, mutation, deformed, blurry, dehydrated, bad anatomy, bad proportions, extra limbs, cloned face, disfigured, gross proportions, malformed limbs, missing arms, missing legs, extra arms, extra legs, fused fingers, too many fingers, long neck"
  - source_model: dreamlike-art/dreamlike-anime-1.0
    positive_prompt: "photo anime, masterpiece, high quality, absurdres, 1girl, 1boy, waifu, chibi"
    negative_prompt: "simple background, duplicate, retro style, low quality, lowest quality, 1980s, 1990s, 2000s, 2005 2006 2007 2008 2009 2010 2011 2012 2013, bad anatomy, bad proportions, extra digits, lowres, username, artist name, error, duplicate, watermark, signature, text, extra digit, fewer digits, worst quality, jpeg artifacts, blurry"
  - source_model: Lykon/dreamshaper-8
    positive_prompt: "bokeh, intricate, elegant, sharp focus, soft lighting, vibrant colors, dreamlike, fantasy, artstation, concept art"
    negative_prompt: "low quality, lowres, jpeg artifacts, signature, bad anatomy, extra legs, extra arms, extra fingers, poorly drawn hands, poorly drawn feet, disfigured, out of frame, tiling, bad art, deformed, mutated, blurry, fuzzy, misshaped, mutant, gross, disgusting, ugly, watermark, watermarks"
  - source_model: dreamlike-art/dreamlike-diffusion-1.0
    positive_prompt: "dreamlikeart, a grungy woman with rainbow hair, travelling between dimensions, dynamic pose, happy, soft eyes and narrow chin, extreme bokeh, dainty figure, long hair straight down, torn kawaii shirt and baggy jeans, In style of by Jordan Grimmer and greg rutkowski, crisp lines and color, complex background, particles, lines, wind, concept art, sharp focus, vivid colors"
    negative_prompt: "nude, naked, low quality, lowres, jpeg artifacts, signature, bad anatomy, extra legs, extra arms, extra fingers, poorly drawn hands, poorly drawn feet, disfigured, out of frame"
\end{lstlisting}
\newpage
\begin{lstlisting}[style=graybox]
Runway ML MoE

base_model: runwayml/stable-diffusion-v1-5
num_experts: 4
moe_layers: all
num_experts_per_tok: 4
type: sd
experts:
  - source_model: runwayml/stable-diffusion-v1-5
    positive_prompt: "ultra realistic, photos, cartoon characters, high quality, anime"
    negative_prompt: "faces, limbs, facial features, in frame, worst quality, hands, drawings, proportions"
  - source_model: CompVis/stable-diffusion-v1-4
    positive_prompt: "ultra realistic, photos, cartoon characters, high quality, anime"
    negative_prompt: "faces, limbs, facial features, in frame, worst quality, hands, drawings, proportions"
  - source_model: CompVis/stable-diffusion-v1-3
    positive_prompt: "ultra realistic, photos, cartoon characters, high quality, anime"
    negative_prompt: "faces, limbs, facial features, in frame, worst quality, hands, drawings, proportions"
  - source_model: CompVis/stable-diffusion-v1-2
    positive_prompt: "ultra realistic, photos, cartoon characters, high quality, anime"
    negative_prompt: "faces, limbs, facial features, in frame, worst quality, hands, drawings, proportions"

\end{lstlisting}

\section{Statistical Analysis} \label{sec:stat_analysis}
For \hyperref[tab:other_scores]{\Cref{tab:other_scores}}, 
\hyperref[tab:mean_length_english_prompts]{\Cref{tab:mean_length_english_prompts}}, and 
\hyperref[tab:mean_length_finnish_prompts]{\Cref{tab:mean_length_finnish_prompts}}, 
we use the Jonckheere-Terpstra test, which is appropriate for detecting ordered trends across multiple groups. In our setting, this tests whether uncertainty quartiles follow a consistent ordering, such as $\mu_1 \geq \mu_2 \geq \mu_3 \geq \mu_4$, rather than merely testing for any difference between groups.

For CLIP Score in \hyperref[tab:other_scores]{\Cref{tab:other_scores}}, the Jonckheere-Terpstra test yields a p-value of $3.34 \times 10^{-19}$, confirming a significant monotonic trend across uncertainty quartiles. To assess practical significance, we also report standardized effect sizes between adjacent quartiles. Using Cohen's $d$, we obtain $d=0.21$ for Q1 vs. Q2, $d=0.88$ for Q2 vs. Q3, and $d=1.18$ for Q3 vs. Q4, corresponding to small, large, and very large effects, respectively.

For English prompt length in \hyperref[tab:mean_length_english_prompts]{\Cref{tab:mean_length_english_prompts}}, the Jonckheere-Terpstra test yields a p-value of $2.13 \times 10^{-12}$ for character count and $1.23 \times 10^{-7}$ for word count. For Finnish prompt length in \hyperref[tab:mean_length_finnish_prompts]{\Cref{tab:mean_length_finnish_prompts}}, the same test yields a p-value of $1.59 \times 10^{-24}$ for character count and $6.79 \times 10^{-22}$ for word count. These results indicate statistically significant trends in which lower-uncertainty prompts are slightly longer on average in both English and Finnish. However, these effects are small and should be interpreted as secondary prompt characteristics rather than the primary driver of EMoE's uncertainty estimates.

Additionally, we use a one-sided t-test to compare Finnish and English uncertainty values, testing $\mu_{\text{Finnish}} > \mu_{\text{English}}$. This test yields a p-value of $9.51 \times 10^{-66}$, confirming that Finnish prompts have significantly higher uncertainty than English prompts. Together, these analyses support the statistical significance of the reported trends while clarifying their practical effect sizes.

\section{Qualitative Results} \label{sec:qualitative_results}
\begin{figure}[h!]
    \centering
    \resizebox{\textwidth}{!}{\input{./figures/paper/uncertain_certain_panel.tex}}
    \caption{EMoE's uncertainty across different prompts: Each row represents a distinct prompt, while the columns denote the output of each component. The left panel displays low uncertainty, while the right panel shows higher uncertainty, indicating more ambiguous or less familiar prompts.}
    \label{fig:uncertainty_panel}
    \vspace{-0.2cm}
\end{figure}

In addition to the examples provided in the main paper, we have included additional qualitative results of our MoE model. \autoref{fig:uncertainty_panel} shows two sets of images: low uncertainty images on the left and high uncertainty images on the right. Each row corresponds to a single prompt, while the columns display the outputs from different ensemble components. The low uncertainty prompts exhibit less variation across ensemble outputs, whereas the high uncertainty prompts show greater diversity, indicating the model’s difficulty in capturing the semantic meaning of the prompt in the generated images. Here, we present another example of models showing bias towards Finnish prompts containing ``pizza'', as illustrated in  \autoref{fig:pizza_bias2}.

\begin{figure}[t!]
    \centering
    \resizebox{\textwidth}{!}{\input{./figures/paper/cold_weather_bias.tex}}
    \caption{Qualitative comparison of image-generation for a Finnish prompt with the word ``pizza" and a random Finnish prompt. Note that the English translation was not provided to the model.}%
    \label{fig:pizza_bias2}
    \vspace{-0.5cm}
\end{figure}

\section{LLM Usage} \label{sec:llm_usage}
This manuscript was written by the authors and later refined for clarity and style using Large Language Models (LLMs).

\end{document}